\RequirePackage{fix-cm}
\documentclass[smallextended]{svjour3} 
\smartqed 
\usepackage{graphicx}
\usepackage{graphics,subfigure}
\usepackage{array}
\usepackage{algorithm}
\usepackage{algorithmic}

\usepackage{booktabs}
\usepackage{microtype}
\usepackage[hyphens]{url}
\usepackage{amsfonts}
\usepackage{amsmath}
\usepackage{amssymb}  
\usepackage{mathptmx} 

\usepackage[misc, geometry]{ifsym}
\usepackage{natbib}
\usepackage{hyperref}
\usepackage{verbatim}
\usepackage{rotating}
\usepackage{bbm}
\usepackage{bm}
\usepackage{cite}
\usepackage{multirow}
\usepackage{calc}
\usepackage{tikz}
\begin{document}

\title{Joint Adaptive Graph and Structured Sparsity Regularization for Unsupervised Feature Selection
}
\subtitle{}


\author{Zhenzhen Sun \and
Yuanlong Yu* 
}


\institute{Z. Sun
\at
College of Computer Science and Technology, HuaQiao University, Xiamen, Fujian, 361021, China. \\
\email{zzsun@hqu.edu.cn} \\
*Corresponding author: Y. Yuan
\at
College of Computer and Data Science, Fuzhou University, Fuzhou, Fujian, 350116, China.\\
\email{yu.yuanlong@fzu.edu.cn} 
}

\date{Received: date / Accepted: date}

\maketitle

\begin{abstract}
Feature selection is an important data preprocessing in data mining and machine learning which can be used to reduce the feature dimension without deteriorating model's performance. Since obtaining annotated data is laborious or even infeasible in many cases, unsupervised feature selection is more practical in reality. Though lots of methods for unsupervised feature selection have been proposed, these methods select features independently, thus it is no guarantee that the group of selected features is optimal. What's more, the number of selected features must be tuned carefully to obtain a satisfactory result. To tackle these problems, we propose a joint adaptive graph and structured sparsity regularization unsupervised feature selection (JASFS) method in this paper, in which a $l_{2,0}$-norm regularization term with respect to transformation matrix is imposed in the manifold learning for feature selection, and a graph regularization term is incorporated into the learning model to learn the local geometric structure of data adaptively. An efficient and simple iterative algorithm is designed to solve the proposed optimization problem with the analysis of computational complexity. After optimized, a subset of optimal features will be selected in group, and the number of selected features will be determined automatically. Experimental results on eight benchmarks demonstrate the effectiveness and efficiency of the proposed method compared with several state-of-the-art approaches.

\end{abstract}

\section{Introduction}
\label{Intro}

Feature selection, a process of selecting a subset of the raw features which are the most relevant and informative, has been widely researched for many years \citep{Liu:18,Zabihimayvan:19,Qasim:20,Bai:20}, and has been used in many real-world applications, \emph{e.g.}, pattern recognition \citep{Xue:21}, text mining \citep{Tang:16,Wan:19}, and bioinformatics \citep{Saeys:07,Lei:20}. Depending on the existing of label guidance, feature selection models can be classified into three categories: supervised, semi-supervised, and unsupervised. Since obtaining annotated data is laborious or even infeasible in many cases, unsupervised feature selection is more challenging while more practical in reality. Therefore, unsupervised feature selection has received more attention currently and tremendous efforts have been made.

In general, the existing unsupervised feature selection methods can be classified into three categories depending on how they combine the feature selection with model learning: filter methods, wrapper methods, and embedded methods. Filter methods \citep{He:05,Qian:13} process feature selection and model learning independently, in which the features are selected according to the intrinsic properties of the data. Wrapper methods \citep{Tabakhi:14} treat the learning algorithm as a black box to score the subsets of features, and embedded methods \citep{Hou:14,Wang:15a,Liu:20} incorporate the feature selection and model learning into a single optimization problem, such that higher computational efficiency and cluster performance can be gained than filter methods and wrapped methods.

Recently, most embedded unsupervised feature selection methods combine the spectral analysis with a structured sparsity regularization term \citep{Gui:17} to select important features, in which a similarity graph is constructed to preserve the manifold structure of original data space. Early spectral-based embedded methods \citep{Hou:14,Yang:11,Li:12,Nie:19,Li:19} use $l_{2,1}$-norm regularization with respect to the model transformation matrix to induce row-sparse solution. After optimization, the importance of individual features is evaluated by the corresponding coefficients and selected one by one. Though satisfactory results can be achieved by using $l_{2,1}$-norm regularization, there are still some limitations. First, using $l_{2,1}$-norm will over-penalize large weights \citep{Qian:15}. An ideal feature selection function should penalize each weight equally to establish a fair competition among all features. Second, selecting features independently will neglect the correlation among features, the selected features are not necessarily to be the optimal feature combination \citep{Qian:15,Du:19,Xiang:13}. Last but not least, using $l_{2,1}$-norm regularization need to tune the regularization parameter as well as the number of selected features. For $l_{2,1}$-norm regularization, even a large value of regularization factor (\emph{e.g.}, $10^5$) cannot produce strong row-sparsity. Thus, the number of selected features also need to be tuned to achieve a satisfactory result after the regularization factor tuned, which will cost a large amount of time and effort and is unpractical. Therefore, it is meaningful to find a way to solve the original $l_{2,0}$-norm regularization problem directly which can produce exact row-sparse solution.

To this end, we propose a novel unsupervised feature selection method, namely, Joint Adaptive Graph and Structured Sparsity Regularization Unsupervised Feature Selection (JASFS), which solves the original $l_{2,0}$-norm regularization problem directly to obtain exact row-sparse solution. Furthermore, a graph regularization term based on maximum entropy \citep{Li:19} is incorporated to learn the similarity matrix adaptively, and a nonnegative orthogonal constrained spectral clustering is used to learn pseudo cluster labels. We design an iterative algorithm to solve the proposed objective function of JASFS, in which an accelerated matrix homotopy iterative hard thresholding (AMHIHT) method is proposed to optimize the transformation matrix in order to reduce the computational time. After optimization, a group of optimal features will be selected and the number of selected features can be determined automatically. To validate the efficiency and effectiveness of the proposed method, we compare it with several recent state-of-the-art unsupervised feature selection methods on eight benchmark datasets, the metrics of cluster accuracy (ACC) and normalized mutual information (NMI) are used for comparison. The results show that the proposed method outperforms the comparison methods.

The main contributions of this paper are listed as follows.
\begin{enumerate}
  \item We propose a Joint Adaptive Graph and Structured Sparsity Regularization Unsupervised Feature Selection (JASFS) method, which perform mainfold learning and feature selection simultaneously to preserve the cluster performance of selected features. Benefited from the $l_{2,0}$-norm regularization, the proposed method can can select features in group determine the number of selected features automatically.
  \item An adaptive graph regularization is incorporated to learn the similarity matrix adaptively, which can reduce the adverse impact of noise and outliers, and thus improve the stability of the model learning.
  \item An efficient iterative algorithm is designed to solve the proposed objective function of JASFS, in which an AMHIHT method is proposed to optimize the transformation matrix in order to reduce the computational time effectively. Experiment results on eight benchmark datasets show the state-of-the-art performance of the proposed method on unsupervised feature selection task.
\end{enumerate}

The rest of this paper is organized as follows. The related unsupervised feature selection works are presented in section \ref{sec:related work}. In Section \ref{sec:methods}, the details of the proposed JASFS is presented, and the computational complexity of the optimization algorithm is analyzed. Section \ref{Sec:Experiment} describes the implementation details and the experimental results. Conclusions and future work are given in Section \ref{Sec:Conclusion}.

\section{Related Work}
\label{sec:related work}

In the past decades, lots of unsupervised feature selection methods have been proposed, these methods can be roughly classified into three categories: filter, wrapper and embedded methods. In this section, we introduce the related embedding methods.

Embedded methods based on spectral regression and sparsity regularization have revised increasing attention in recent years. Cai \emph{et al.} proposed a multiple cluster feature selection (MCFS) method in \citep{Cai:10}. MCFS first use graph Lapalcian to construct pseudo cluster labels for input data, then a sparse regression based on $l_1$-norm is used to calculate the relevant coefficients between the pseudo cluster labels and features. Finally, all the features are sorted according to their MCFS scores in descending order and those of the highest scores are selected. Unsupervised discriminative feature selection (UDFS) \citep{Yang:11} incorporates discriminative analysis and $l_{2,1}$-norm minimization into a joint framework for unsupervised feature selection. Nonnegative discriminative feature selection (NDFS) \citep{Li:12} imposes a nonnegative constraint with respect to cluster labels into the objective function to exploit discriminative information for feature selection. Robust unsupervised feature selection (RUFS) \citep{Qian:13} uses local learning regularized robust nonnegative matrix factorization to learn pseudo cluster labels, which is more robust to redundant or noisy features. Robust spectral feature selection (RSFS) \citep{Shi:14} improves the robustness of graph embedding and sparse spectral regression jointly. Subsequently, Nie \emph{et al}. \citep{Nie:19} proposed a structure optimal graph feature selection (SOGFS) method, which performs feature selection and local structure learning simultaneously, the similarity matrix thus can be determined adaptively. Meanwhile, Li \emph{et al}. \citep{Li:19} proposed an uncorrelated regression with adaptive graph for unsupervised feature selection (URAFS), which imposes a generalized uncorrelated constraint to seek the uncorrelated yet discriminative features. What's more, a graph regularized term based on maximum entropy is incorporated into the model to learn local geometric structure adaptively. Later, Wang \emph{et al}. \citep{Wang:21} proposed an unsupervised soft-label feature selection (USFS) model, which performs soft-label learning and simultaneously guides the unsupervised feature selection process with the learned soft-labels to improve the cluster performance of selected features.

Most of above methods use $l_{2,1}$-norm regularization for feature selection, which will suffer from some problems as mentioned in section \ref{Intro}. The most relevant work to our approach are adaptive unsupervised feature selection (AUFS) \citep{Qian:15} and unsupervised group feature selection (UGFS) \citep{Du:19}, all of which use $l_{2,0}$-norm as the regularization term for feature selection instead of $l_{2,1}$-norm. AUFS uses a joint adaptive loss for data fitting and a $l_{2,0}$-norm regularization for feature selection, and optimize the transformation matrix by  proximal gradient descent method, which is more time-consumption than our optimization method. UGFS tries to solve a $l_{2,0}$-norm equality constraint problem, in which the number of selected features must be predefined to construct the equality constraint. How many features need be selected is unknown, when tuning the number of selected features, the optimization program must be run repeatedly, which will cost a large amount of time and effort and is unpractical. What's more, this two methods construct similarity matrix and select features independently, the similarity matrix is calculated from original data and remains as constant for the following procedure, but in real applications data always contain lots of noise, which make the similarity matrix unreliable \citep{Wang:15}. As pointed out in \citep{Nie:19}, the unreliable similarity matrix will damage the local manifold structure, and ultimately lead to suboptimal result. Different from these two methods, we combine the spectral regression with the $l_{2,0}$-norm regularization for feature selection, which can determine the number of selected features automatically, and a graph regularization term is used to learn the local geometric structure of data adaptively

\section{Methodology}
\label{sec:methods}

In this section, we first introduce the notations and defines used in this paper. Then, the framework of the proposed method is presented and an iterative optimization approach is provided to solve the objective function. Finally, we analyze the computational complexity of the optimization approach and combined with some related works.

\subsection{Notations and Definitions}
The notations and definitions used in this paper are presented in this section. The vector are represented by boldface lowercase letters, and matrices are represented as boldface uppercase letters. For a vector $ \boldsymbol{x} \in R^n $, $x_i$ denotes its $i-th$ element. For a matrix $\boldsymbol{X}=\{x_{ij}\}  \in R^{n\times m} $, $ \boldsymbol{x}^{i} $ and $ \boldsymbol{x}_{j}$ denote its $i-th$ row and $j-th$ column, respectively. $tr(\boldsymbol{X})$ denotes the trace of matrix $\boldsymbol{X}$, and $\boldsymbol{X}^T$ denotes its transpose. $\textbf{1}=[1,1,...,1]^T \in R^{n}$ is the vector with all elements $1$, and $\boldsymbol{I}$ is the identity matrix. A centering matrix is defined as $\boldsymbol{H} = \boldsymbol{I} - (1/n)\textbf{1}\textbf{1}^T$.

For $p \ne 0$, the $p$-norm of the vector $ \boldsymbol{x}$ is defined as
\begin{equation}\notag
||\boldsymbol{x}||_p=(\sum_{i=1}^n{|x_i|^p}) ^{\text{1/}p}.
\end{equation}

The Frobenius norm of matrix $ \boldsymbol{X} $ is defined as
\begin{equation}\notag
||\boldsymbol{X}||_F = \ \sqrt{\sum_{i = 1}^n{\sum_{j = 1}^m{x_{ij}^2}}} = \ \sqrt{tr(\boldsymbol{X}^T \boldsymbol{X})}. \\
\end{equation}

The $l_{2,1}$-norm of $ \boldsymbol{X} $ is defined as
\begin{equation}\notag
||\boldsymbol{X}||_{2,1} =  \ \sum_{i =1}^n{||\boldsymbol{x}^i||_2} = \ \sum_{i =1}^n{\sqrt{\sum_{j = 1}^m{x_{ij}^2}}}. \\
\end{equation}

The $l_{2,0}$-norm of $\boldsymbol{X}$ is defined as
\begin{equation}\notag
||\boldsymbol{X}||_{2,0} =  \ \sum_{i =1}^n{ \mathbbm{1}_{||\boldsymbol{x}^{i}||_2 \ne 0}}, \\
\end{equation}
where $ \mathbbm{1}_{A} $ stands for the indicator function. For a scalar $x$, if $x \ne 0$, $\mathbbm{1}_{x} = 1$, otherwise $\mathbbm{1}_{x} = 0$. Thus the $l_{2,0}$-norm of matrix $ \boldsymbol{X} $ is defined as the number of non-zero rows in $\boldsymbol{X} $. If a matrix has a large number of zero rows, we define it has the property of row-sparsity.

\subsection{Nonnegative Spectral Learning}

In unsupervised learning paradigm, spectral theory has been successfully applied for many years \citep{Nie:11,Li:20}, due to it can discover the local geometrical structure of the original data without label guidance. Li \emph{et al}. \citep{Li:12} proposed a nonnegative spectral learning model to learn the pseudo-label more precise for unlabeled data. For a set of unlabeled data $\boldsymbol{X}=\{\boldsymbol{x}_1, \boldsymbol{x}_2,..., \boldsymbol{x}_N\} \in R^{d \times N}$, where $\boldsymbol{x}_i$ is the $i-th$ sample and the data samples are drawn from $C$ clusters. Nonnegative spectral learning solves the following problem:
\begin{align}\label{NSL}
  \underset{\boldsymbol{W,F,b}}{\min}\,\,\frac{1}{2}||\boldsymbol{W}^T\boldsymbol{X}+\boldsymbol{b}\boldsymbol{1}^T-\boldsymbol{F}||_{F}^{2}+\alpha tr\left(\boldsymbol{F}\boldsymbol{L}_S\boldsymbol{F}^T \right)
\notag \\
\,\,  s.t.  \ \boldsymbol{F}\boldsymbol{F}^T=\boldsymbol{I}, \boldsymbol{F}\geq \text{0}.  \ \ \ \ \ \ \ \ \ \ \ \ \ \ \ \ \ \ \
\end{align}
where $\boldsymbol{W}\in R^{d\times c}$ is the full rank transformation matrix, $\boldsymbol{b}\in R^{C\times 1}$ is the bias. $\boldsymbol{F} = \{\boldsymbol{f}_1, \boldsymbol{f}_2,...,\boldsymbol{f}_N\} \in R^{C \times N}$ denotes the pseudo-label indicator matrix embedded in the
input data space, and $\boldsymbol{f}_i$ is the label indicator of $i-th$ sample. $\boldsymbol{L}_S = \boldsymbol{D} - (\boldsymbol{S}^T + \boldsymbol{S})/2$ is the Laplacian graph related to the similarity matrix $\boldsymbol{S} =\{s_{ij}\} \in R^{N \times N}$, and $\boldsymbol{D}$ is the degree matrix calculated as
\begin{equation}\label{degreematix}
  \boldsymbol{D}=\text{diag}\left( \sum_{j=1}^N{\frac{s_{1j}+s_{j1}}{2},..,\sum_{j=1}^N{\frac{s_{Nj}+s_{jN}}{2}}} \right).
\end{equation}

\subsection{Adaptive Graph Learning}

In spectral learning, the similarity matrix $\boldsymbol{S}$ is crucial for the final performance. For traditional methods, $\boldsymbol{S}$ is always constructed by KNN and computed based on Euclidean distance before model leaning, that is, $\boldsymbol{S}$ is set as constant in model \eqref{NSL}. However, the original data always contain a large number of noise and outliers that make the similarity matrix created by original data cannot be fully relied. In order to address this problem, researchers have made a lot of efforts to learn the similarity matrix adaptively in recent years \citep{Nie:10,Wang:18,Yu:21}. Nie \emph{et al}. \citep{Nie:19} first proposed a structured optimal graph for feature selection with a $l_2$-norm minimization constrained on $\boldsymbol{S}$. Subsequently, Li \emph{et al}. \citep{Li:19} proposed an adaptive graph regularization for feature selection based on the theory of maximum entropy. In this paper, we applied the adaptive graph regularization to construct the similarity matrix. That is, the similarity matrix should be optimized by
\begin{equation}\label{AGR}
  \underset{\sum_{j=1}^N{s_{ij}=1,s_{ij}\geqslant 0}}{\max}\,\,\sum_{i=1}^N{\sum_{j=1}^N{\left( -s_{ij}\log s_{ij} \right)}}.
\end{equation}
For problem \eqref{AGR}, the optimal solution is $s_{ij} = \frac{1}{N}(\forall i,j)$, which reflects the most reliable state when there is no prior knowledge about data.

\subsection{Feature Selection}
In our framework, we employ the $l_{2,0}$-norm regularization constrained on the transformation matrix to achieve feature selection, in which a row-sparsity solution can be obtained and the features corresponding to non-zero rows will be selected. Combined with \eqref{NSL} and \eqref{AGR}, the overall objective function of JASFS can be derived as
\begin{align}\label{JASFS}
\underset{\boldsymbol{W,F,b,S}}{\min}\,\,\frac{1}{2}||\boldsymbol{W}^T\boldsymbol{X}+\boldsymbol{b}\mathbf{1}^T-\boldsymbol{F||}_{F}^{2}+\alpha tr\left(\boldsymbol{F}\boldsymbol{L}_S\boldsymbol{F}^T  \right) +\beta \sum_{i=1}^N{\sum_{j=1}^N{s_{ij}\log s_{ij}}}+\lambda ||\boldsymbol{W||}_{\text{2,}0}
\notag\\
\,\,\,s.t. \boldsymbol{FF}^T=\boldsymbol{I,F}\geq \text{0,}\sum_{j=1}^N{s_{ij}}=\text{1, }s_{ij}\geq \text{0}. \ \ \ \ \ \ \ \ \ \ \ \ \ \ \ \ \ \ \ \ \ \ \ \ \ \ \ \ \ \ \ \ \ \ \ \ \
\end{align}
where $\alpha, \beta, \lambda \in R_{+}$ are regularization coefficients.

According to Karush-Kuhn-Tucker (KKT) theorem \citep{Karush:14}, set the derivative of \eqref{JASFS} with respect to $\boldsymbol{b}$ to $0$, then we can obtain the optimal solution of $\boldsymbol{b}$ as follows:
\begin{equation}\label{solutionofb}
  \boldsymbol{b}=\frac{1}{N}\left(\boldsymbol{F}-\boldsymbol{W}^T\boldsymbol{X} \right) \mathbf{1}.
\end{equation}
Replacing $\boldsymbol{b}$ with equation~\eqref{solutionofb}, we can rewrite problem \eqref{JASFS} as
\begin{align}\label{JASFS1}
 \underset{\boldsymbol{W,F,S}}{\min}\,\,\frac{1}{2}||(\boldsymbol{W}^T\boldsymbol{X}-\boldsymbol{F})\boldsymbol{H}||_{F}^{2}+2\alpha tr\left(\boldsymbol{F}\boldsymbol{L}_S\boldsymbol{F}^T  \right) +\beta \sum_{i=1}^N{\sum_{j=1}^N{s_{ij}\log s_{ij}}}+\lambda ||\boldsymbol{W}||_{2,0}
\notag \\
\,\,s.t. \ \boldsymbol{F}\boldsymbol{F}^T=\boldsymbol{I, F}\geq \text{0, }\sum_{j=1}^N{s_{ij}}=\text{1, }s_{ij}\geq 0. \ \ \ \ \ \ \ \ \ \ \ \ \ \ \ \ \ \ \ \ \ \ \ \ \
\end{align}
Now, the objective function of JASFS is presented totally, next we will show how to optimize it.

\subsection{Optimization Procedure}
An alternative optimization algorithm is derived to solve problem \eqref{JASFS1}.

(1) \emph{Update $\boldsymbol{W}$ with fixed $\boldsymbol{F}$ and $\boldsymbol{S}$}

The objective function with respect to $\boldsymbol{W}$ is
\begin{equation}\label{objofW}
 \varphi(\boldsymbol{W}) = \underset{\boldsymbol{W}}{\min}\,\,\frac{1}{2}||(\boldsymbol{W}^T\boldsymbol{X}-\boldsymbol{F})\boldsymbol{H}||_{F}^{2}+\lambda ||\boldsymbol{W}||_{2,0},
\end{equation}
we have proposed an accelerated matrix homotopy iterative hard thresholding (AMHIHT) to solve this problem in \citep{Sun:22}, which can be used to update $\boldsymbol{W}$ in this paper directly, where $\boldsymbol{W}$ has a closed-form solution at each iteration as
\begin{align}\label{eq:iht_solution}
(\boldsymbol{w}^i)^{t+1}  = \left\{ {\begin{array}{*{20}c}
	{(\boldsymbol{w}^i)^{t} - \frac{1}{{L}}\nabla f((\boldsymbol{w}^i)^{t}),if\;||(\boldsymbol{w}^i)^{t}  - \frac{\nabla f((\boldsymbol{w}^i)^{t})}{{L}}||_2^2  > \frac{{2\lambda }}{{L}}}  \\
	{\textbf{0},\;\;otherwise\;\;\;\;\;\;\;\;\;\;\;\;\;\;\;\;\;\;\;\;\;\;\;\;\;\;\;\;}  \\
	\end{array}} \right.
\end{align}
where $\nabla f(\boldsymbol{W})$ is the gradient of $f(\boldsymbol{W}) = 1/2 ||(\boldsymbol{W}^T\boldsymbol{X}-\boldsymbol{F})\boldsymbol{H}||_{F}^{2}$, and $ L \ge 0 $ is an upper bound on the Lipschitz constant of $\nabla f(\boldsymbol{W})$, i.e., $L \geq L_f$. The procedure of AMHIHT method is described in Algorithm \ref{Alg:AMHIHT}, and the convergence analysis of Algorithm \ref{Alg:AMHIHT} can be found in \citep{Sun:22}.
\begin{algorithm}[t]
\caption{Acceleration Matrix Homotopy Iterative Hard Thresholding Algorithm for solving problem \eqref{objofW}}
\label{Alg:AMHIHT}
\begin{algorithmic}
  \STATE \text{\textbf{(Input:)} Training data $\textbf{X} \in R^{d \times N}$, cluster labels $\textbf{F} \in R^{C \times N}$, centering matrix $\textbf{H} \in R^{N \times N}, \textbf{W}^0$;}
  \STATE \text{parameters $\lambda, L_0,{\lambda}_0,L_{min},L_{max};// L_0 \in [L_{min},L_{max}]$;}
  \STATE \text{\textbf{(Output:)} \ $\boldsymbol{W}^*$;}
  \STATE \text{1: initialize $\rho \in (0,1),\gamma >1, \eta >0$, set $k \gets 0$;} \\
          \text{2: \textbf{repeat}}\\
          \text{ \ \ \emph{An L-tuning iteration}}\\
          \text{3: \ \ update $\boldsymbol{W}^{k+1}$ by Eq.~\eqref{eq:iht_solution};}\\
          \text{4: \ \ \textbf{while} \ $ {\varphi}_{\lambda _k}(\boldsymbol{W}^{k})- {\varphi}_{\lambda _k}(\boldsymbol{W}^{k+1})< \frac{\eta}{2} ||\boldsymbol{W}^{k}-\boldsymbol{W}^{k+1}||_F^2$ \ \textbf{do}}\\
          \text{5: \ \ \  $L_{k} \gets min \{ \gamma L_{k},L_{max} \}$;}\\
          \text{6: \ \ \  update $\boldsymbol{W}^{k+1}$ by Eq.~\eqref{eq:iht_solution};}\\
          \text{7: \ \ \textbf{end while}}\\
          \text{8: \ \  $L_{k+1} \gets L_{k}$;}\\
          \text{9: \ \ ${\lambda}_{k+1} \gets \rho {\lambda}_k$;}\\
          \text{10: \ \ $k \gets k+1$;}\\
          \text{11: \textbf{until} ${\lambda}_{k+1} \leq \lambda$}\\
          \text{12: $\boldsymbol{W}^{\star} \gets \boldsymbol{W}^k$.}
\end{algorithmic}
\end{algorithm}

(2) \emph{Update $\boldsymbol{F}$ with fixed $\boldsymbol{W}$ and $\boldsymbol{S}$}

The objective function with respect to $\boldsymbol{F}$ is
\begin{align}\label{objofF}
 \underset{\boldsymbol{F}}{\min}\,\,\frac{1}{2}||(\boldsymbol{W}^T\boldsymbol{X}-\boldsymbol{F})\boldsymbol{H}||_{F}^{2}
 +\alpha tr\left(\boldsymbol{F}\boldsymbol{L}_S\boldsymbol{F}^T \right)  \ \ \ \ \
\notag \\
\,\,s.t. \ \boldsymbol{F}\boldsymbol{F}^T=\boldsymbol{I, F}\geq \text{0 }. \ \ \ \ \ \ \ \ \ \ \ \ \ \ \ \
\end{align}

Fist, we rewrite problem \eqref{objofF} as follows:
\begin{align}\label{objofF1}
 \underset{\boldsymbol{F}}{\min}\,\,\frac{1}{2}||(\boldsymbol{W}^T\boldsymbol{X}-\boldsymbol{F})\boldsymbol{H}||_{F}^{2}
 +\alpha tr\left(\boldsymbol{F}\boldsymbol{L}_S\boldsymbol{F}^T \right) + \frac{\nu}{2}||\boldsymbol{F}\boldsymbol{F}^T-\boldsymbol{I||}_{F}^{2}
\notag \\
 s.t. \ \boldsymbol{F}\geq \text{0 }. \ \ \ \ \ \ \ \ \ \ \ \ \ \ \ \ \ \ \ \ \ \ \ \ \ \ \ \ \ \
\end{align}
where $\nu > 0$ is a parameter to control the orthogonality condition. In practice, $\nu$ should be large enough to insure the orthogonality satisfied.

Denotes
\begin{equation}\label{PQ}
  \boldsymbol{P} = \frac{1}{2}\boldsymbol{H}+\alpha\boldsymbol{L}_S, \ \ \boldsymbol{Q}= \frac{1}{2}\boldsymbol{W}^T\boldsymbol{X}\boldsymbol{H},
\end{equation}
problem \eqref{objofF1} can be rewritten as
\begin{align}\label{objofF2}
 \underset{\boldsymbol{F}}{\min}\,\, tr\left(\boldsymbol{F}\boldsymbol{PF}^T -2\boldsymbol{F}\boldsymbol{Q}^T \right) + \frac{\nu}{2}||\boldsymbol{F}\boldsymbol{F}^T-\boldsymbol{I||}_{F}^{2}
\notag \\
 \,\,s.t. \ \boldsymbol {F}\geq \text{0 }. \ \ \ \ \ \ \ \ \ \ \ \ \ \ \ \ \ \ \ \ \ \ \ \
\end{align}
the Lagrange function of \eqref{objofF2} is
\begin{equation}\label{Lagrange}
   tr\left(\boldsymbol{F}\boldsymbol{PF}^T -2\boldsymbol{F}\boldsymbol{Q}^T \right) + \frac{\nu}{2}||\boldsymbol{F}\boldsymbol{F}^T-\boldsymbol{I||}_{F}^{2} + tr(\boldsymbol{\varPhi}\boldsymbol{F}^T)
\end{equation}
where $\boldsymbol{\varPhi}=\left\{\phi _{ij}\right\}$ are the Lagrange multipliers for constrain $F_{ij}\geq 0$.

According to the KKT condition, set the derivative of \eqref{objofF2} with respect to $F_{ij}$ to $0$ and $\phi_{ij}F_{ij} = 0$, the updating rule for $\boldsymbol{F}$ is obtained as
\begin{equation}\label{slutonofF}
 F_{ij}\,\,\gets \,\,F_{ij}\frac{\left( \nu \boldsymbol{F}+\boldsymbol{Q} \right) _{ij}}{\left(\boldsymbol{FP}+\nu \boldsymbol{FF}^T\boldsymbol{F} \right) _{ij}}.
\end{equation}
then, we normalize $\boldsymbol{F}$ such that $(\boldsymbol{F}\boldsymbol{F}^T)_{ii} =1, i=1,...,C$.

(3) \emph{Update $\boldsymbol{S}$ with fixed $\boldsymbol{W}$ and $\boldsymbol{F}$}

With $\boldsymbol{W}$ and $\boldsymbol{F}$ fixed, the objective function \eqref{JASFS1} is transformed into
\begin{align}\label{objofS}
 \underset{\boldsymbol{S}}{\min}\,\, \alpha tr\left(\boldsymbol{F}\boldsymbol{L}_S\boldsymbol{F}^T \right) + \beta \sum_{i=1}^N{\sum_{j=1}^N{s_{ij}\log s_{ij}}}  \ \ \ \ \
\notag \\
\,\,s.t. \ \sum_{j=1}^N{s_{ij}}=\text{1, }s_{ij}\geq 0. \ \ \ \ \ \ \ \ \ \ \
\end{align}
According to \citep{Li:19}, the optimal $s_{ij}$ can be computed as
\begin{equation}\label{solutionofS}
s_{ij}=\frac{\exp \left( -\frac{\alpha||\boldsymbol{f}_i-\boldsymbol{f}_j||_{2}^{2}}{2\beta} \right)}{\sum_{j=1}^N{\exp \left( -\frac{||\alpha\boldsymbol{f}_i-\boldsymbol{f}_j||_{2}^{2}}{2\beta} \right)}}
\end{equation}

With the alternative optimization of $\boldsymbol{W}$, $\boldsymbol{F}$ and $\boldsymbol{S}$, the iterative optimization of proposed JASFS \eqref{JASFS} is summarized in Algorithm \ref{Alg:JASFS}.
\begin{algorithm}[t]
\caption{Optimization Algorithm of JASFS \eqref{JASFS}}
\label{Alg:JASFS}
\begin{algorithmic}
  \STATE \text{\textbf{(Input:)} Training data $\boldsymbol{X} \in R^{d \times N}$, clustering number $C$, parameters $\lambda, \alpha, \beta, \nu$;}\\
  \STATE \text{1: initialize $k \gets 0, \boldsymbol{W}^0=\textbf{0},$ construct $\boldsymbol{F}^0$ by \emph{K-means}, compute $\boldsymbol{S}^0$ by Eq.~\eqref{solutionofS};}\\
  \STATE  \text{2: \textbf{repeat}}\\
          \text{3: \ \ update $\boldsymbol{W}^{k+1}$ by Alg.~\ref{Alg:AMHIHT};}\\
          \text{4: \ \ Calculate $\boldsymbol{L}_S$ with $\boldsymbol{L}_S=\boldsymbol{D}-\frac{\boldsymbol{S}^T+\boldsymbol{S}}{2}$, calculate $\boldsymbol{P}$ and $\boldsymbol{Q}$ with Eq.~\eqref{PQ};}\\
          \text{5: \ \ Update $\boldsymbol{F}^{k+1}$ by Eq.~\eqref{slutonofF};}\\
          \text{6: \ \ Update $\boldsymbol{S}^{k+1}$ by Eq.~\eqref{solutionofS};}\\
          \text{7: \ \ $k \gets k+1$;}\\
          \text{8: \textbf{until} convergence}\\
          \text{9: \textbf{Output:} Selected features corresponding to non-zero rows of $\textbf{W}$.}
\end{algorithmic}
\end{algorithm}

\subsection{Computational Complexity Analysis}
In this subsection, the computational complexity of the proposed optimization algorithm is analysed. As show in Algorithm \ref{Alg:JASFS}, the problem \eqref{JASFS1} is optimized by addressing subproblems \eqref{objofW}, \eqref{objofF1}, and \eqref{objofS} iteratively. In subproblem \eqref{objofW}, the most time-consuming step is to calculate the gradient $\nabla f\left(\boldsymbol{W} \right)$, which has the complexity of $\mathcal{O}\left( dNC \right)$. The most computational steps of \eqref{objofF1} and \eqref{objofS} are $\mathcal{O}\left( dN^{2} \right)$ and $\mathcal{O}\left(N^2 \right)$. Therefore, the computational complexity of JASFS is $\mathcal{O}\left(dNC + dN^{2} + N^2 \right)$. Table \ref{Complexity} illustrates the computational complexity of current state-of-the-art methods, in which $m$ is the reduced dimension in related methods, and $p$ is the number of neighbors in graph construction. From this table it can be seen that, for high-dimensional data, our approach, RUFS, AUFS and UGFS are much efficient than other methods, since the computational complexity of most methods are depend on $d^3$, while these four methods are only depend on $d$.
\begin{table}[htp]
\caption{Computational Complexity Comparison}\label{Complexity}
\centering
\begin{tabular}{c|c}
\hline\noalign{\smallskip}
{Method}&{Computational Complexity}\\
\noalign{\smallskip}\hline\noalign{\smallskip}
{MCFS}&{$\mathcal{O}\left(d^3 + N^{2}m + d^{2}N \right)$}\\
{UDFS}&{$\mathcal{O}\left(d^3 + N^{2}C \right)$}\\
{NDFS}&{$\mathcal{O}\left(d^3 + N^{2}C \right)$}\\
{RUFS}&{$\mathcal{O}\left(dNC + N^{2}C \right)$}\\
{AUFS}&{$\mathcal{O}\left(dNC + N^{2}C + NC \right)$}\\
{SOGFS}&{$\mathcal{O}\left(d^3 + N^{2}p + N^2C \right)$}\\
{URAFS}&{$\mathcal{O}\left(d^3 + dN^{2} + N^2 \right)$}\\
{USFS}&{$\mathcal{O}\left(d^2N^{2} + dm + NC + dC + dNC \right)$}\\
{UGFS}&{$\mathcal{O}\left(dNC + dN^{2} + CN^2 + C^2N^2 + NlogN \right)$}\\
{JASFS}&{$\mathcal{O}\left(dNC + dN^{2} + N^2 \right)$}\\
\noalign{\smallskip}\hline
\end{tabular}
\end{table}
\section{Experiments}
\label{Sec:Experiment}

In this section, we conduct extensive experiments to verify the efficiency and effectiveness of the proposed JASFS. The experiments include the comparison of performance in clustering task, the comparison of running time, and the parameter sensitivity of the proposed method. We use eight benchmark datasets to evaluate the proposed method, including Brain \citep{Pomeroy:02}, Breast3 \citep{Veer:02}, Jaffe \citep{Lyons:98}, Lung \citep{Bhattacharjee:01}, Mnist \citep{LeCun:98}, NCI \citep{Jeffrey:02}, ORL \citep{Cai:10} and Palm \citep{Zhang:03}. Table \ref{Datasets} shows a detail introduction to these data sets. All data and code used to perform these experiments are available online\footnote{https://github.com/zhenzhenSun-FZU/JASFS} to help with reproducibility.

\begin{table}[htp]
\caption{Datasets Desciption}\label{Datasets}
\centering
\begin{tabular}{cccc}
\hline\noalign{\smallskip}
{Datasets}&{\#samples}&{\#Features}&{\#Classes}\\
\noalign{\smallskip}\hline\noalign{\smallskip}
{Brain}&{42}&{5597}&{5}\\
{Breast3}&{95}&{4869}&{3}\\
{Jaffe}&{213}&{676}&{10}\\
{Lung}&{203}&{3312}&{5}\\
{Mnist}&{4000}&{784}&{10}\\
{NCI}&{61}&{5244}&{8}\\
{ORL}&{400}&{1024}&{40}\\
{Palm}&{2000}&{4096}&{100}\\
\noalign{\smallskip}\hline
\end{tabular}
\end{table}
\subsection{Experimental Setup}
We compare our method with the Baseline (all features) and nine state-of-the-art unsupervised feature selection methods: L-score \citep{He:05}, MCFS \citep{Cai:10}, UDFS \citep{Yang:11}, NDFS \citep{Li:12}, RUFS \citep{Qian:13}, AUFS \citep{Qian:15}, URAFS \citep{Li:19}, USFS \citep{Wang:21}and UGFS \citep{Du:19}, these methods are described in section \ref{sec:related work} in detail.

We search the hyper-parameters in the grid of $\{10^{-6}, 10^{-4}, 10^{-2}, 1, 10^{2}, 10^{4},$ $10^{6}\}$ for all methods except $\lambda$ of JASFS, which is search in $\{10^{-6}, 10^{-5}, 10^{-4},$ $10^{-3}, 10^{-2}, 10^{-1}\}$, because if $\lambda \geq 1$, the optimal solution for $\boldsymbol{W}$ is zero. In JASFS, $\nu$ is set as $10^8$ to guarantee the orthogonality satisfied as suggested in \citep{Li:12}. For our method, the features corresponding to non-zero rows of $\boldsymbol{W}$ are selected, then we record the number of selected features (No.fea). As most methods set, the number of selected features is tuned from $\{50, 100,...,400\}$ for all compared methods. For all methods, the initial cluster labels matrix $\boldsymbol{F}$ is constructed by \emph{K-means}, in which the number of neighbors is set as $5$. The other parameters are set the same values as the authors designed. After feature selection, the \emph{K-means} clustering algorithm is applied on the selected features to predict cluster labels for testing, and the value $k$ of \emph{K-means} is set as the number of classes. Since the result of \emph{K-means} is sensitive to initialization, ten repeated trials are carried out with the same random seeds for each method, and then we record the average results for comparison and analysis. The methods with higher average and lower variances are regarded as more accuracy ones. We also record corresponding No.fea with best result for comparison.

All experiments are implemented by MATLAB2014a in a standard PC with hardware configuration as follows:
\begin{enumerate}
  \item CPU: Intel(R) Pentium(R) CPU G2030 @3.40GHz;
  \item Memory: 32.00GB;
  \item Graphics Processing Unit (GPU): None.
\end{enumerate}

\subsection{Validation Metric}
Two metrics are used in the experiments: clustering accuracy (ACC) \citep{Papadimitriou:82} and normalized mutual information (NMI) \citep{Fan:49}. ACC is defined as
\begin{equation}\label{ACC}
 \text{ACC}=\frac{1}{N}\sum_{i=1}^N{\delta \left( t_i,\text{map}\left( c_i \right) \right)},
\end{equation}
where $t_i$ is the ground truth of the $i-th$ sample, and $c_i$ is the computed clustering label by \emph{K-means}. $\delta(.)$ denotes the $\delta-$function, where $\delta(x,y) = 1$ if $x=y$ and $0$ otherwise. map$(.)$ denotes the best mapping function that maps cluster labels to match the ground truth using Kuhn-Munkres algorithm \citep{Strehl:02}.

NMI is defined as
\begin{equation}\label{NMI}
  \text{NMI}=\frac{\sum_{k=1}^C{\sum_{m=1}^C{n_{k,m}}\log \frac{N\cdot n_{k,m}}{n_k\hat{n}_m}}}{\sqrt{\left( \sum_{k=1}^C{n_k}\log \frac{n_k}{N} \right) \left( \sum_{m=1}^C{\hat{n}_m}\log \frac{\hat{n}_m}{N} \right)}}
\end{equation}
where $n_k$ denotes the number of samples contained in the cluster $k$ $(1 \leq k \leq C)$, and ${\hat{n}_m}$ is the number of samples of ground truth class $m$ $(1 \leq m \leq C)$, $n_{k,m}$ represents the amount of samples in the intersection between computed cluster $k$ and ground truth class $m$. NMI is in the range of $[0,1]$ and a larger NMI indicates a better clustering result.

\subsection{Clustering Results with Selected Features}
\begin{figure}[htp]
  \begin{center}
    \subfigure[]{\label{Fig:AUFS}
        \includegraphics[width=4.5cm,height=3.7cm]{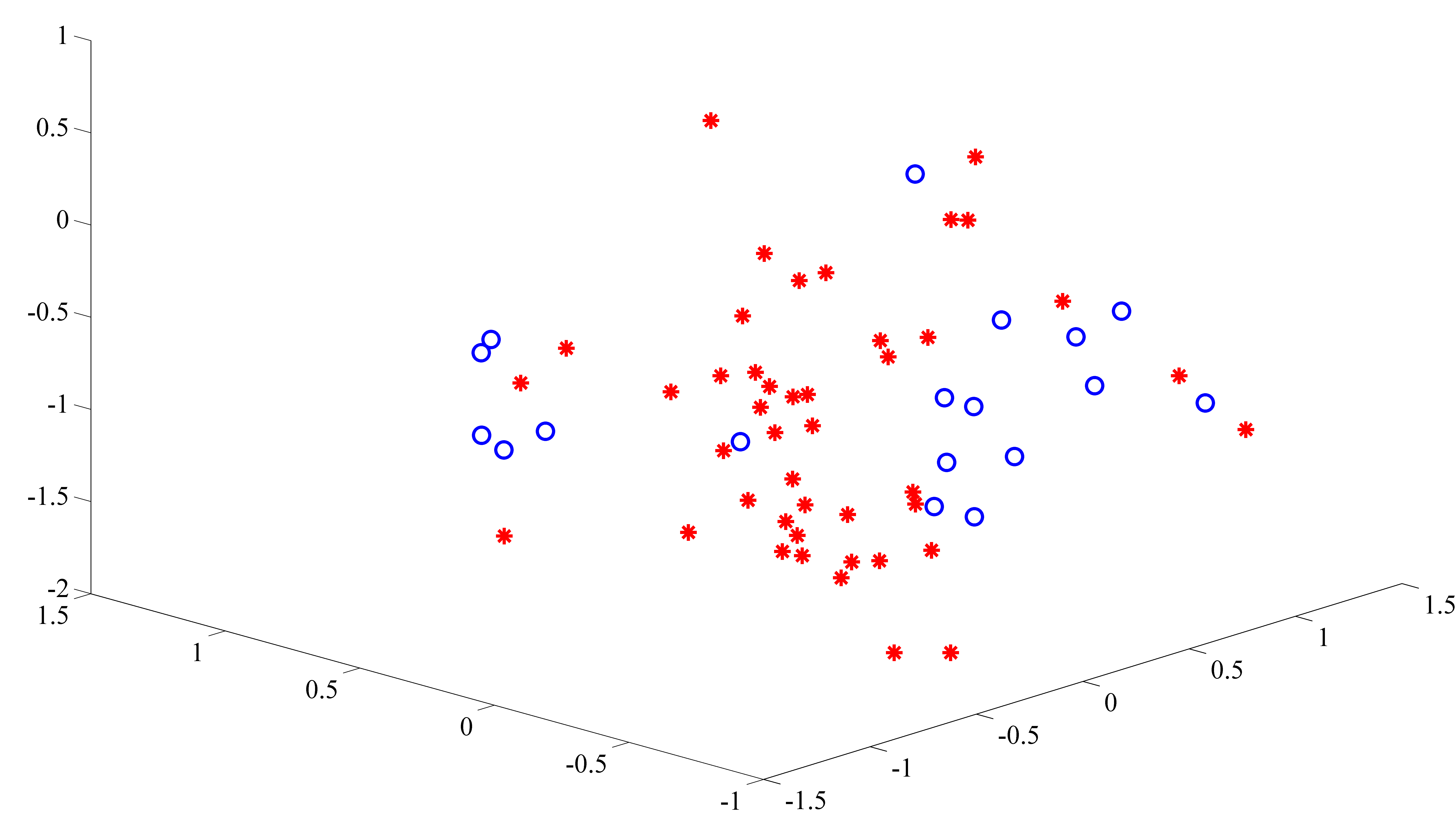}
    }
    \subfigure[]{\label{Fig:UGFS}
        \includegraphics[width=4.5cm,height=3.7cm]{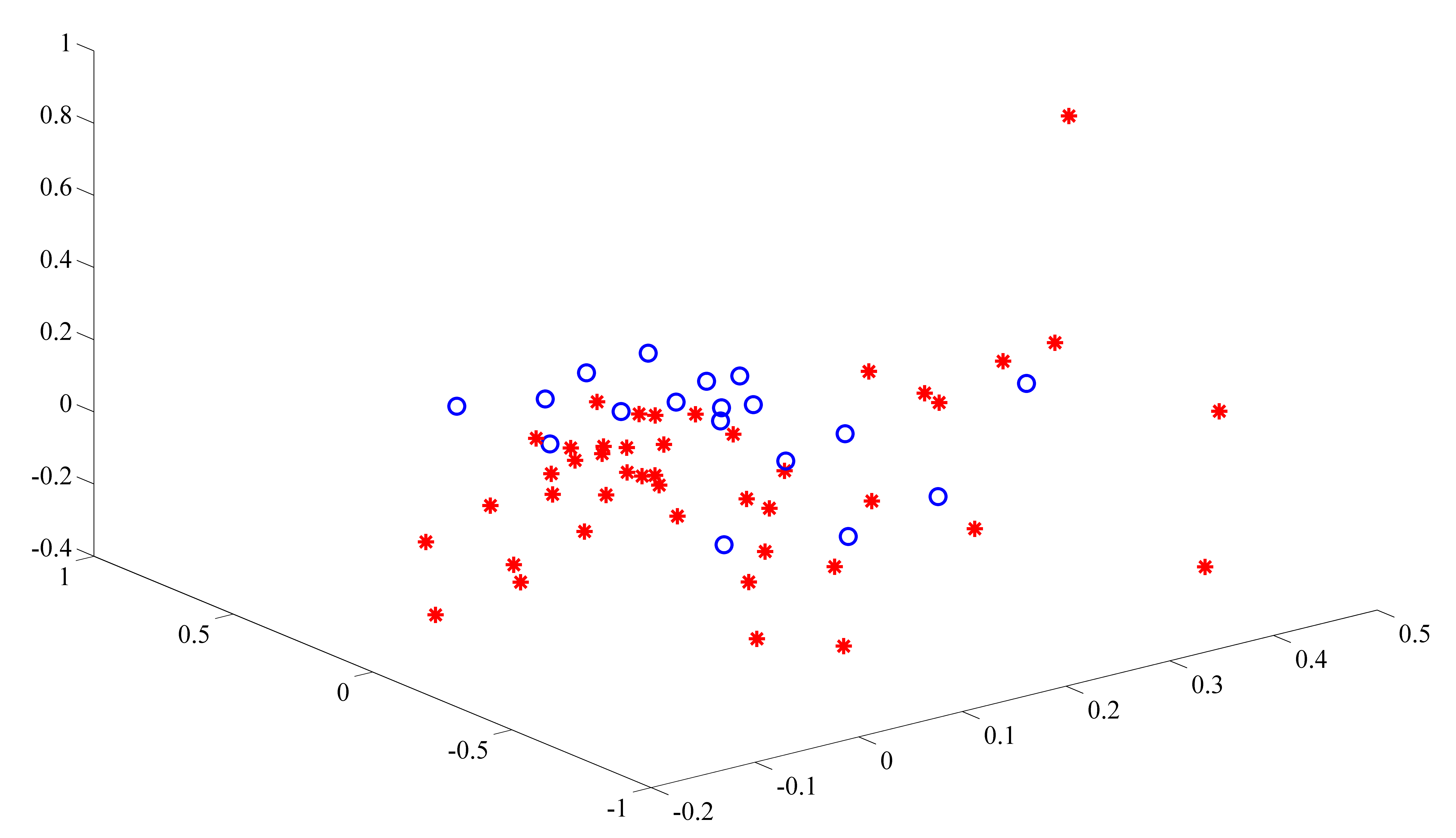}
    }
    \subfigure[]{\label{Fig:JASFS}
        \includegraphics[width=4.5cm,height=3.7cm]{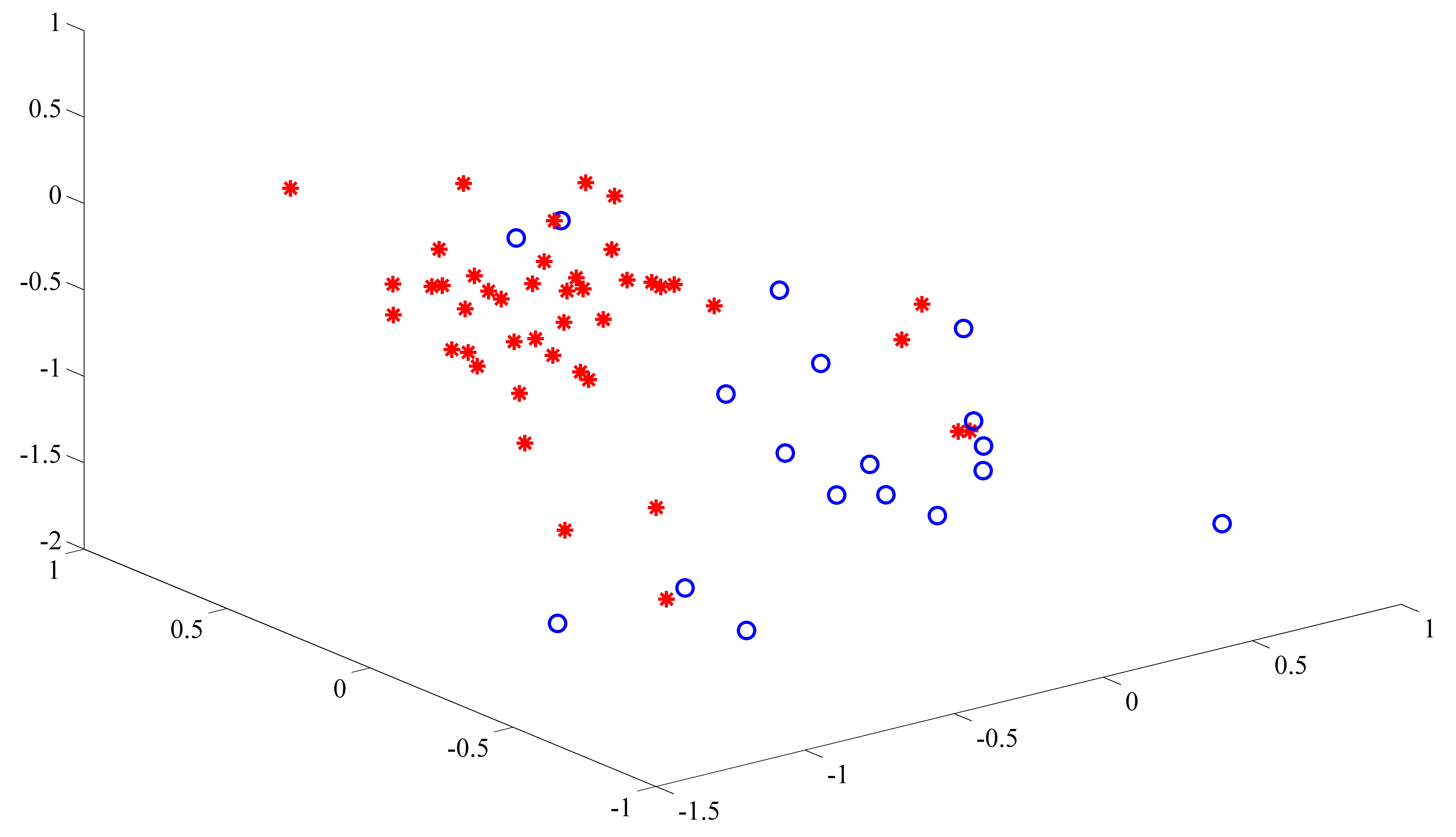}
    }
    \caption{\label{fig:exampleofselectedfea}The Breast3 data points (extracts from two clusters) constituted by the most three important features selected by AUFS, UGFS and JASFS, respectively.
      \subref{Fig:AUFS}: AUFS.
      \subref{Fig:UGFS}: UGFS.
      \subref{Fig:JASFS}: JASFS.
      }
  \end{center}
\end{figure}

The comparison clustering results are divided into two parts. First, the best results obtained by each method are compared, which are shown in Table \ref{tabofacc} and Table \ref{tabofnmi}, Table \ref{Nofea} records the corresponding number of selected features. In the perspective of ACC, our method beats other compared methods on most datasets, which achieves an improvement of more than $2\%$ over the second best result on Brain and Lung datasets. Compared to the Baseline, it can be found that most methods can beat the Baseline on all datasets, demonstrating that feature selection can improve the clustering accuracy. From the perspective of NMI, our approach can achieve best result on most datasets, and the result on Mnist are very close to the best one. The standard deviation results show that the features selected by our approach are much more stable for clustering than other methods. From Table \ref{Nofea} we can see that, the number of selected features used by JASFS is much less than other methods for most datasets. This result implies that JASFS can obtain better results when use fewer features. From the results, we can conclude that JASFS outperforms other methods and can get a better balance between ACC and NMI. Figure \ref{fig:exampleofselectedfea} shows an example that the three most important features selected by AUFS, UGFS and JASFS, respectively. It can be seen that the features selected by our approach are more discriminative for clustering than that selected by the other two $l_{2,0}$-norm based methods. Therefore, our approach can achieve better clustering performance.
\begin{table}[htp]
\caption{Clustering ACC of different unsupervised feature selection approaches.}\label{tabofacc}
\centering
\begin{tabular}{p{1.5cm}<{\centering}|p{1.6cm}<{\centering}p{1.6cm}<{\centering}p{1.6cm}<{\centering}p{1.6cm}<{\centering}p{1.6cm}<{\centering}p{1.6cm}<{\centering}p{1.6cm}<{\centering}p{1.6cm}<{\centering}}
\hline\noalign{\smallskip}
{Method}&{Brain}&{Breast3}&{Jaffe}&{Lung}&{Mnist}&{Nci}&{ORL}&{Palm}\\
\noalign{\smallskip}\hline\noalign{\smallskip}
{Baseline}&{60.24$\pm$8.48}&{54.95$\pm$3.05}&{71.22$\pm$8.80}&{64.78$\pm$8.91}&{51.52$\pm$4.22}&{55.25$\pm$7.08}&{50.73$\pm$3.30}&{68.94$\pm$1.81}\\
{L-score}&{70.48$\pm$8.46}&{62.95$\pm$2.53}&{73.66$\pm$9.96}&{77.93$\pm$9.33}&{52.80$\pm$3.30}&{59.51$\pm$6.48}&{48.13$\pm$2.46}&{60.60$\pm$2.02}\\
{MCFS}&{69.76$\pm$7.07}&{58.95$\pm$3.12}&{77.93$\pm$11.42}&{77.59$\pm$10.65}&{51.91$\pm$3.27}&{60.33$\pm$8.09}&{52.55$\pm$3.50}&{70.38$\pm$2.43}\\
{UDFS}&{65.00$\pm$5.39}&{58.11$\pm$1.78}&{78.92$\pm$8.55}&{71.58$\pm$9.96}&{54.20$\pm$3.83}&{59.18$\pm$5.43}&{52.78$\pm$4.96}&{69.55$\pm$2.42}\\
{NDFS}&{72.14$\pm$6.64}&{60.00$\pm$2.67}&{79.30$\pm$9.78}&{77.93$\pm$5.81}&{54.43$\pm$2.88}&{62.79$\pm$5.47}&{54.80$\pm$3.42}&{71.25$\pm$2.01}\\
{RUFS}&{71.19$\pm$9.35}&{60.42$\pm$2.44}&{73.38$\pm$4.57}&{74.58$\pm$10.63}&{\textbf{56.78$\pm$3.21}}&{61.48$\pm$4.90}&{53.45$\pm$3.49}&{\textbf{72.22$\pm$1.70}}\\
{AUFS}&{71.19$\pm$5.88}&{59.68$\pm$4.00}&{75.54$\pm$8.50}&{80.54$\pm$4.81}&{51.48$\pm$3.26}&{56.39$\pm$6.70}&{46.87$\pm$3.25}&{68.04$\pm$2.46}\\
{URAFS}&{68.57$\pm$10.31}&{59.26$\pm$2.45}&{73.19$\pm$7.52}&{71.38$\pm$8.10}&{51.64$\pm$3.01}&{58.85$\pm$2.00}&{49.07$\pm$3.10}&{65.57$\pm$1.27}\\
{USFS}&{70.00$\pm$6.75}&{61.05$\pm$4.05}&{83.62$\pm$8.07}&{75.37$\pm$8.07}&{54.64$\pm$1.16}&{61.64$\pm$4.10}&{50.63$\pm$2.51}&{71.40$\pm$2.49}\\
{UGFS}&{65.95$\pm$5.03}&{59.26$\pm$5.37}&{77.09$\pm$6.59}&{69.21$\pm$7.36}&{54.92$\pm$2.93}&{58.52$\pm$5.24}&{51.70$\pm$3.19}&{71.54$\pm$3.89}\\
{JASFS}&{\textbf{74.29$\pm$5.24}}&{\textbf{63.37$\pm$1.20}}&{\textbf{84.46$\pm$9.49}}&{\textbf{82.91$\pm$5.45}}&{55.01$\pm$2.02}&{\textbf{63.77$\pm$5.91}}&{\textbf{55.25$\pm$3.01}}&{70.07$\pm$2.31}\\
\noalign{\smallskip}\hline
\end{tabular}
\end{table}

\begin{table}[htp]
\caption{Clustering NMI of different unsupervised feature selection approaches.}\label{tabofnmi}
\centering
\begin{tabular}{p{1.5cm}<{\centering}|p{1.6cm}<{\centering}p{1.6cm}<{\centering}p{1.6cm}<{\centering}p{1.6cm}<{\centering}p{1.6cm}<{\centering}p{1.6cm}<{\centering}p{1.6cm}<{\centering}p{1.6cm}<{\centering}}
\hline\noalign{\smallskip}
{Method}&{Brain}&{Breast3}&{Jaffe}&{Lung}&{Mnist}&{Nci}&{ORL}&{Palm}\\
\noalign{\smallskip}\hline\noalign{\smallskip}
{Baseline}&{48.89$\pm$9.22}&{20.87$\pm$2.26}&{81.86$\pm$4.71}&{58.97$\pm$5.50}&{46.08$\pm$2.03}&{56.34$\pm$5.65}&{73.69$\pm$1.20}&{90.51$\pm$0.53}\\
{L-score}&{59.73$\pm$5.23}&{24.09$\pm$1.93}&{84.27$\pm$4.52}&{63.36$\pm$6.57}&{46.03$\pm$1.74}&{61.42$\pm$3.05}&{71.58$\pm$1.56}&{85.12$\pm$0.61}\\
{MCFS}&{61.11$\pm$5.94}&{21.60$\pm$1.94}&{85.19$\pm$6.08}&{67.16$\pm$5.85}&{46.95$\pm$0.98}&{61.15$\pm$6.21}&{74.23$\pm$2.31}&{90.43$\pm$0.64}\\
{UDFS}&{52.80$\pm$5.01}&{21.10$\pm$1.75}&{76.19$\pm$2.92}&{60.89$\pm$6.60}&{47.04$\pm$2.16}&{62.14$\pm$3.08}&{74.32$\pm$1.65}&{89.36$\pm$0.84}\\
{NDFS}&{62.56$\pm$6.32}&{22.02$\pm$1.72}&{84.42$\pm$4.48}&{61.71$\pm$2.25}&{49.46$\pm$1.66}&{61.74$\pm$4.77}&{75.32$\pm$1.39}&{90.86$\pm$0.70}\\
{RUFS}&{63.51$\pm$8.90}&{\textbf{24.16$\pm$3.43}}&{76.19$\pm$2.92}&{60.89$\pm$6.60}&{48.59$\pm$1.55}&{62.14$\pm$3.08}&{74.32$\pm$1.65}&{\textbf{91.14$\pm$0.54}}\\
{AUFS}&{61.47$\pm$6.26}&{21.45$\pm$2.38}&{80.33$\pm$3.31}&{64.67$\pm$3.95}&{45.72$\pm$1.29}&{58.76$\pm$5.06}&{69.71$\pm$1.15}&{89.16$\pm$0.86}\\
{URAFS}&{56.41$\pm$11.70}&{21.30$\pm$2.29}&{78.46$\pm$4.33}&{57.42$\pm$4.98}&{45.33$\pm$1.31}&{59.94$\pm$3.32}&{72.19$\pm$2.56}&{86.83$\pm$0.54}\\
{USFS}&{57.29$\pm$6.65}&{21.94$\pm$1.41}&{86.60$\pm$4.98}&{64.35$\pm$2.88}&{\textbf{49.89$\pm$1.22}}&{61.23$\pm$3.38}&{71.85$\pm$1.57}&{90.72$\pm$0.62}\\
{UGFS}&{54.79$\pm$8.84}&{22.67$\pm$2.45}&{84.90$\pm$4.24}&{55.32$\pm$3.36}&{48.84$\pm$1.89}&{58.95$\pm$5.59}&{73.57$\pm$1.58}&{91.13$\pm$0.88}\\
{JASFS}&{\textbf{64.99$\pm$4.97}}&{22.88$\pm$1.49}&{\textbf{88.18$\pm$3.11}}&{\textbf{67.79$\pm$3.25}}&{49.76$\pm$0.58}&{\textbf{63.34$\pm$2.93}}&{\textbf{75.72$\pm$1.30}}&{89.55$\pm$0.41}\\
\noalign{\smallskip}\hline
\end{tabular}
\end{table}

\begin{table}[htp]
\caption{Corresponding No.fea that each method achieved its own best result}\label{Nofea}
\centering
\begin{tabular}{c|cccccccc}
\hline\noalign{\smallskip}
{Method}&{Brain}&{Breast3}&{Jaffe}&{Lung}&{Mnist}&{Nci}&{ORL}&{Palm}\\
\noalign{\smallskip}\hline\noalign{\smallskip}
{L-score}&{300}&{150}&{250}&{400}&{400}&{250}&{300}&{400}\\
{MCFS}&{350}&{150}&{150}&{100}&{200}&{350}&{100}&{200}\\
{UDFS}&{150}&{150}&{50}&{300}&{400}&{300}&{250}&{100}\\
{NDFS}&{150}&{350}&{100}&{200}&{250}&{250}&{200}&{300}\\
{RUFS}&{300}&{50}&{400}&{400}&{350}&{100}&{400}&{200}\\
{AUFS}&{250}&{50}&{100}&{200}&{400}&{400}&{350}&{400}\\
{URAFS}&{350}&{50}&{200}&{200}&{400}&{300}&{350}&{250}\\
{USFS}&{400}&{50}&{350}&{350}&{350}&{350}&{300}&{300}\\
{UGFS}&{250}&{100}&{150}&{250}&{400}&{150}&{400}&{300}\\
{JASFS}&{80}&{10}&{240}&{182}&{150}&{88}&{267}&{200}\\
\noalign{\smallskip}\hline
\end{tabular}
\end{table}

\begin{figure}[htp]
  \begin{center}
    \subfigure[]{\label{Fig:Brain}
        \includegraphics[width=4cm,height=3.5cm]{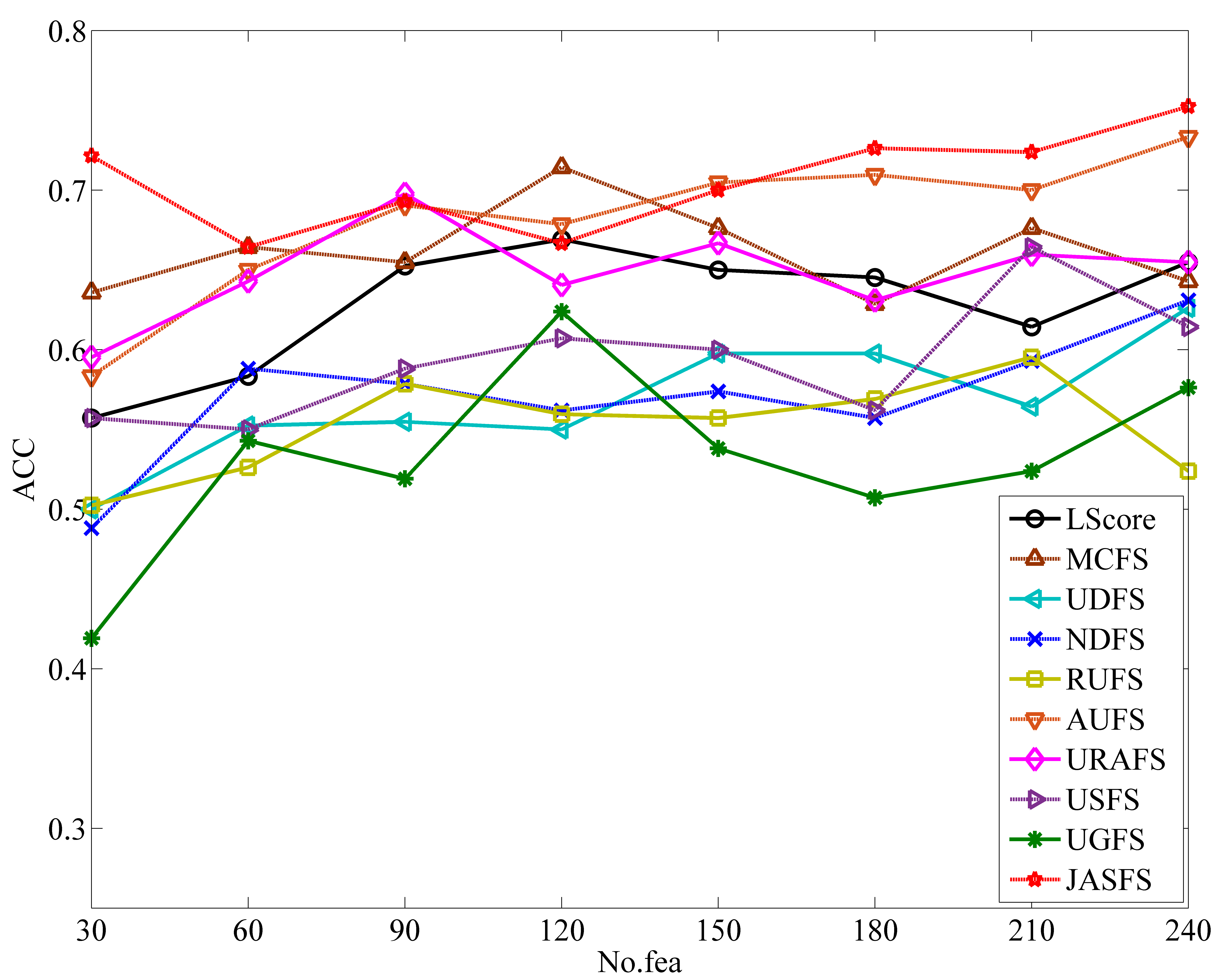}
    }
    \subfigure[]{\label{Fig:Breast3}
        \includegraphics[width=4cm,height=3.5cm]{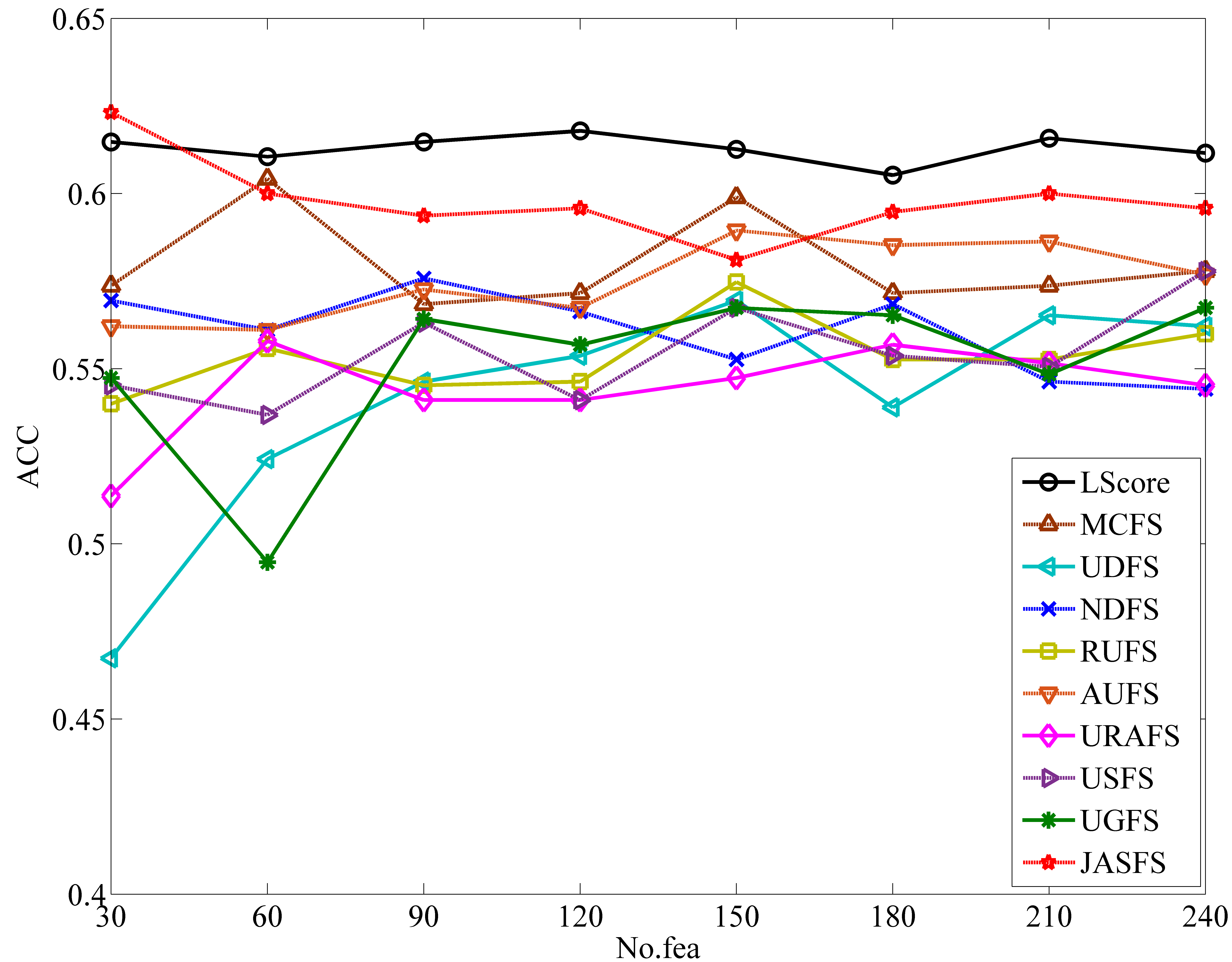}
    }
    \subfigure[]{\label{Fig:Jaffe}
        \includegraphics[width=4cm,height=3.5cm]{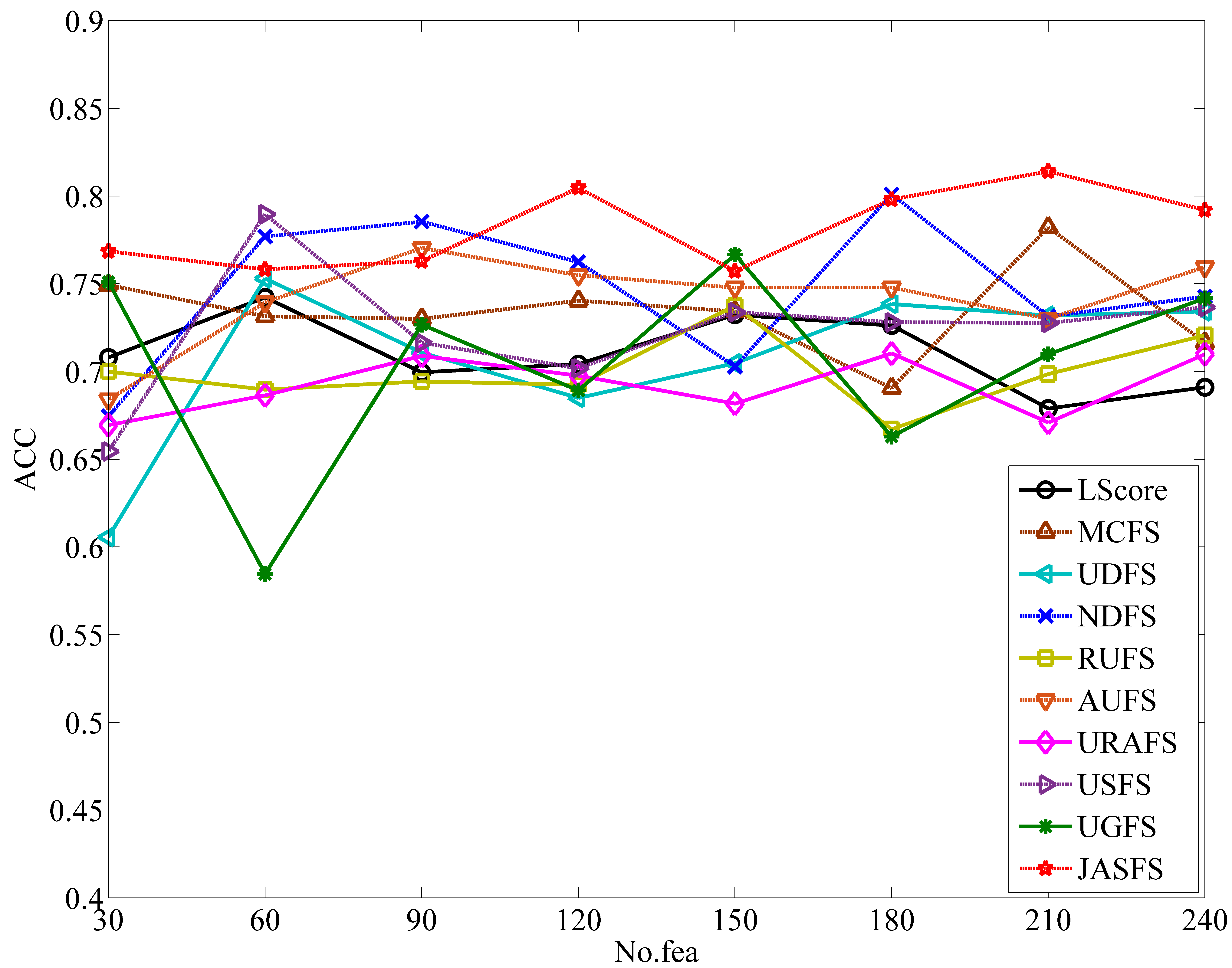}
    }
    \subfigure[]{\label{Fig:Lung}
        \includegraphics[width=4cm,height=3.5cm]{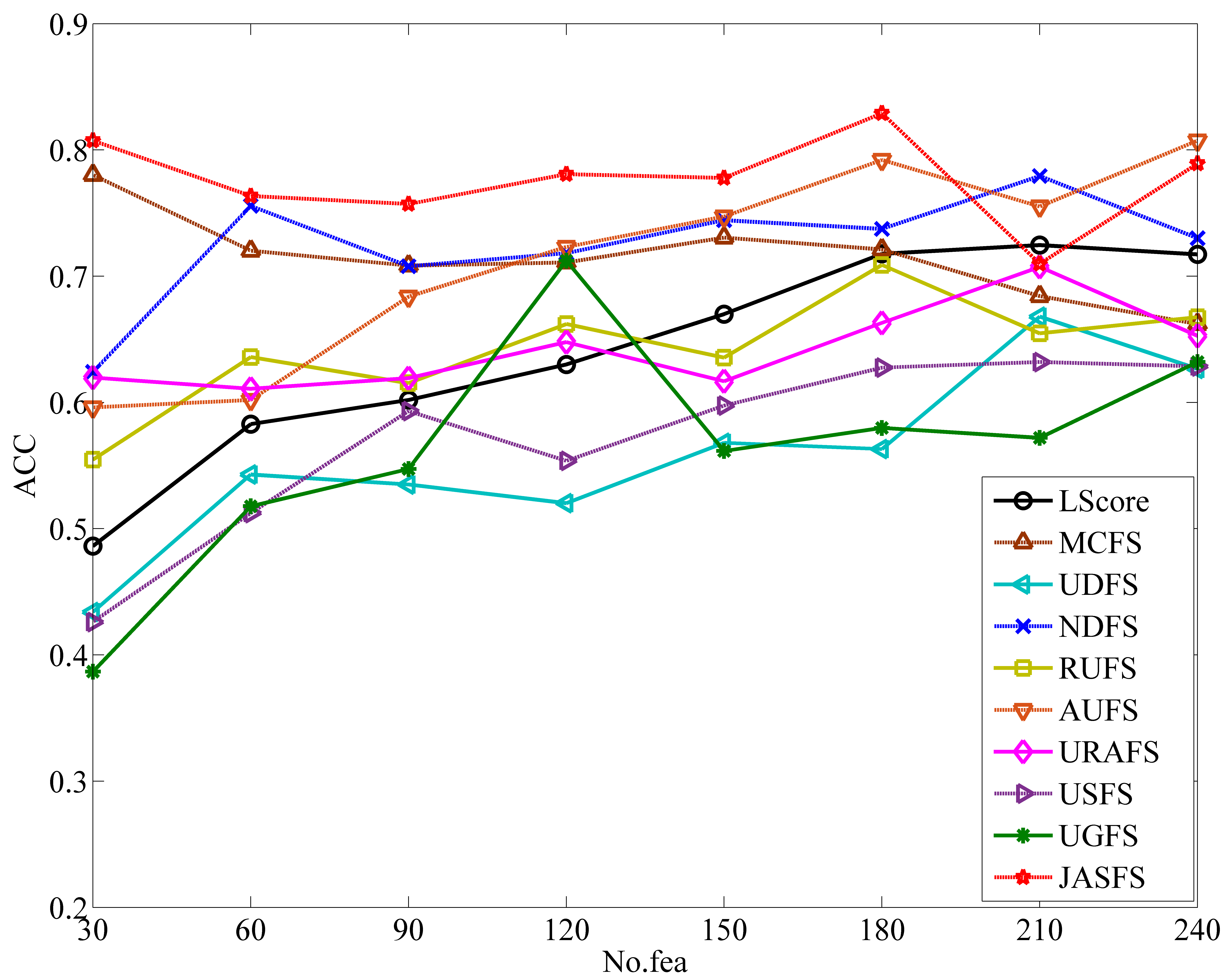}
    }
    \subfigure[]{\label{Fig:Mnist}
        \includegraphics[width=4cm,height=3.5cm]{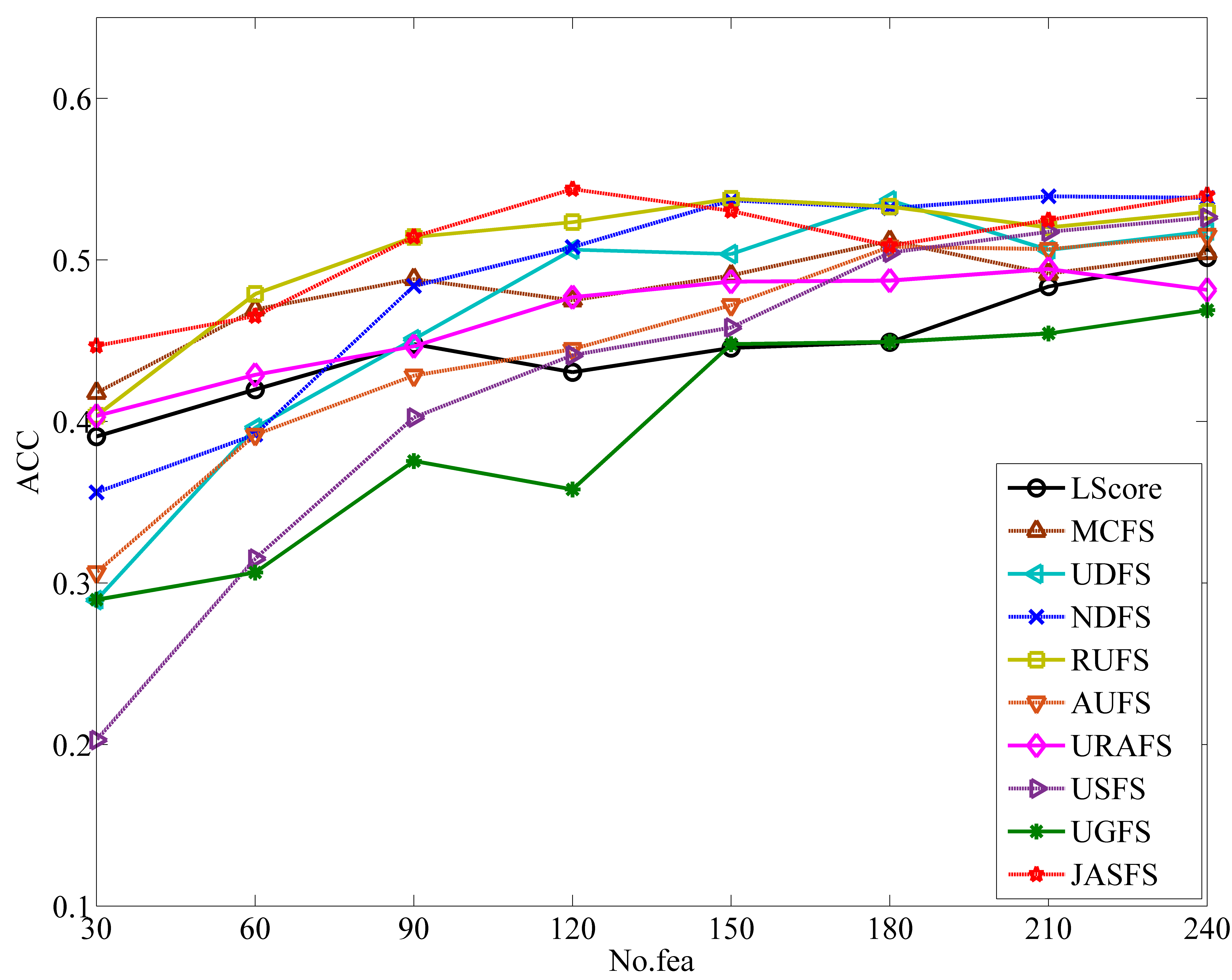}
    }
    \subfigure[]{\label{Fig:NCI}
        \includegraphics[width=4cm,height=3.5cm]{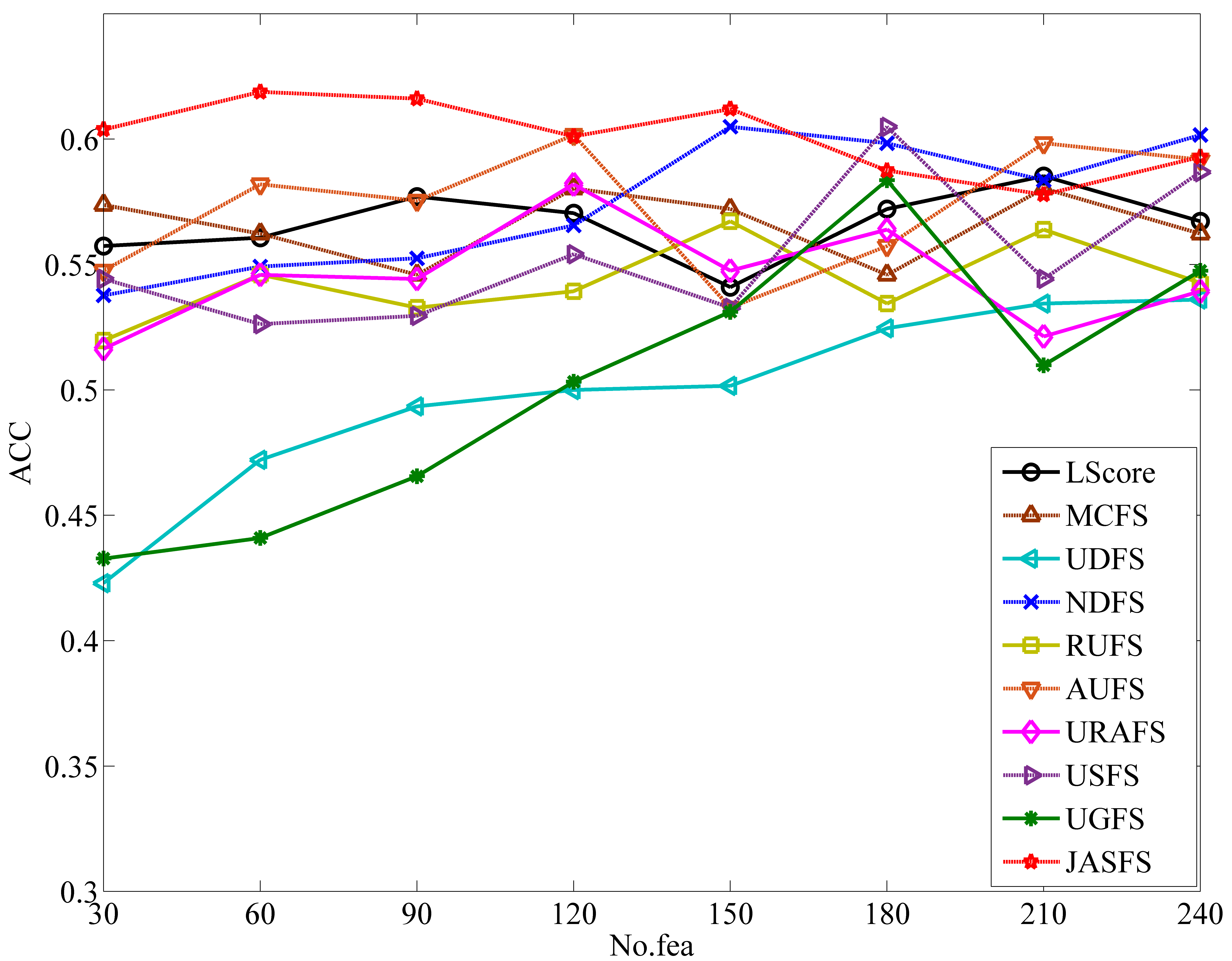}
    }
    \subfigure[]{\label{Fig:ORL}
        \includegraphics[width=4cm,height=3.5cm]{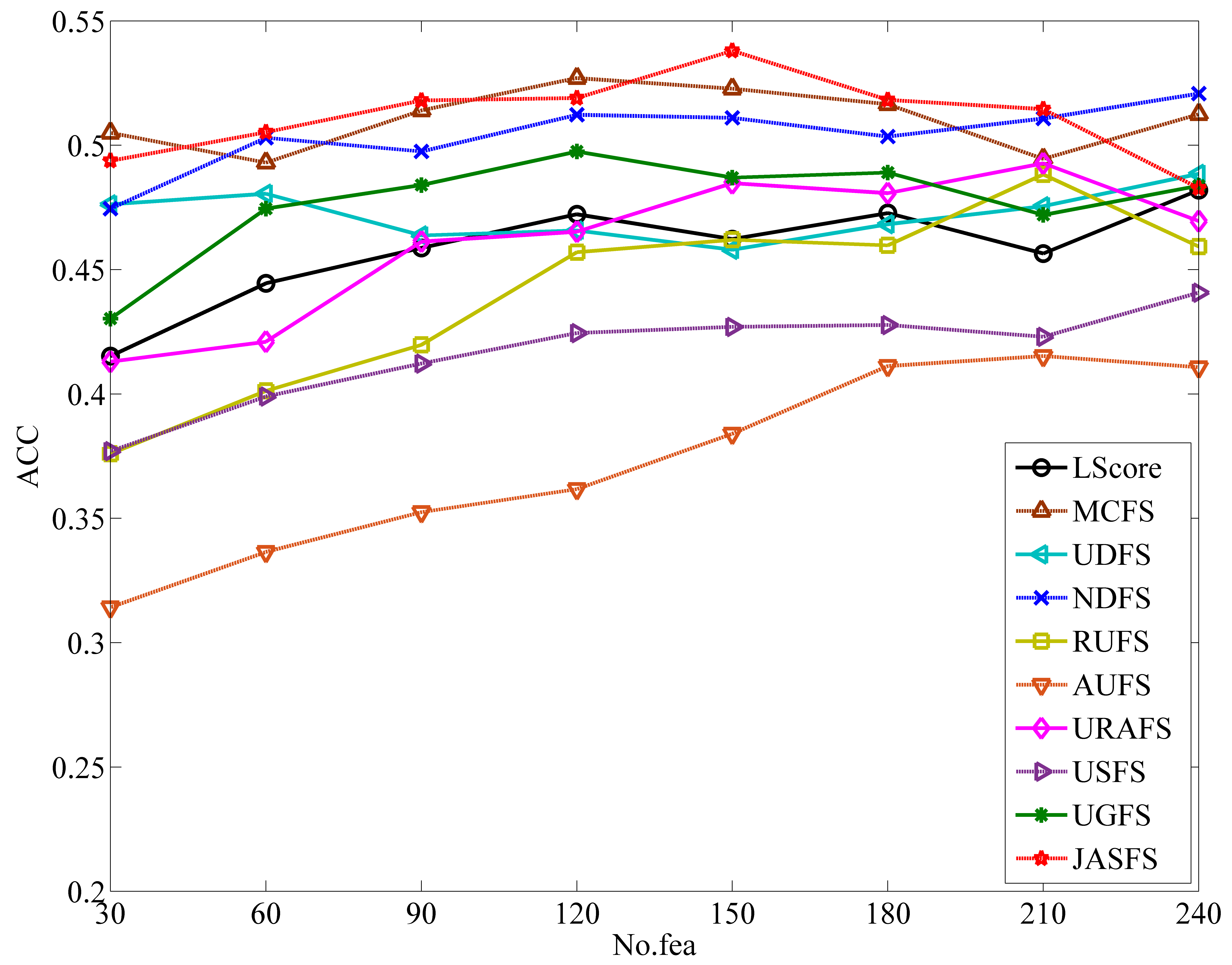}
    }
    \subfigure[]{\label{Fig:Palm}
        \includegraphics[width=4cm,height=3.5cm]{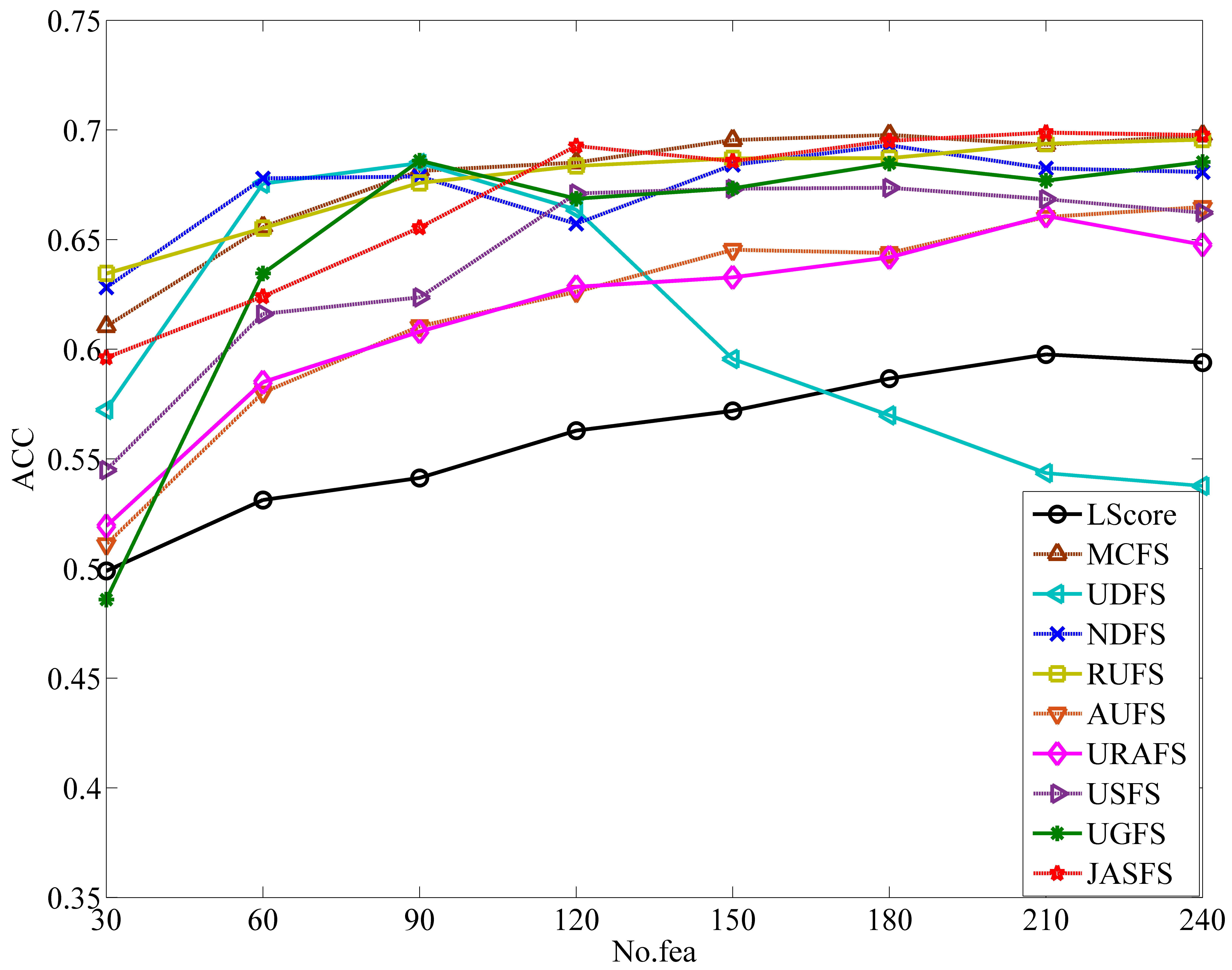}
    }
    \caption{\label{fig:accvsnum}The ACC results with respect to different numbers of selected features.
      \subref{Fig:Brain}: Brain.
      \subref{Fig:Breast3}: Breast3.
      \subref{Fig:Jaffe}: Jaffe.
      \subref{Fig:Lung}: Lung.
      \subref{Fig:Mnist}: Mnist.
      \subref{Fig:NCI}: NCI.
      \subref{Fig:ORL}: ORL.
      \subref{Fig:Palm}: Palm.
      }
  \end{center}
\end{figure}

\begin{figure}[htp]
  \begin{center}
    \subfigure[]{\label{Fig:Brain}
        \includegraphics[width=4cm,height=3.5cm]{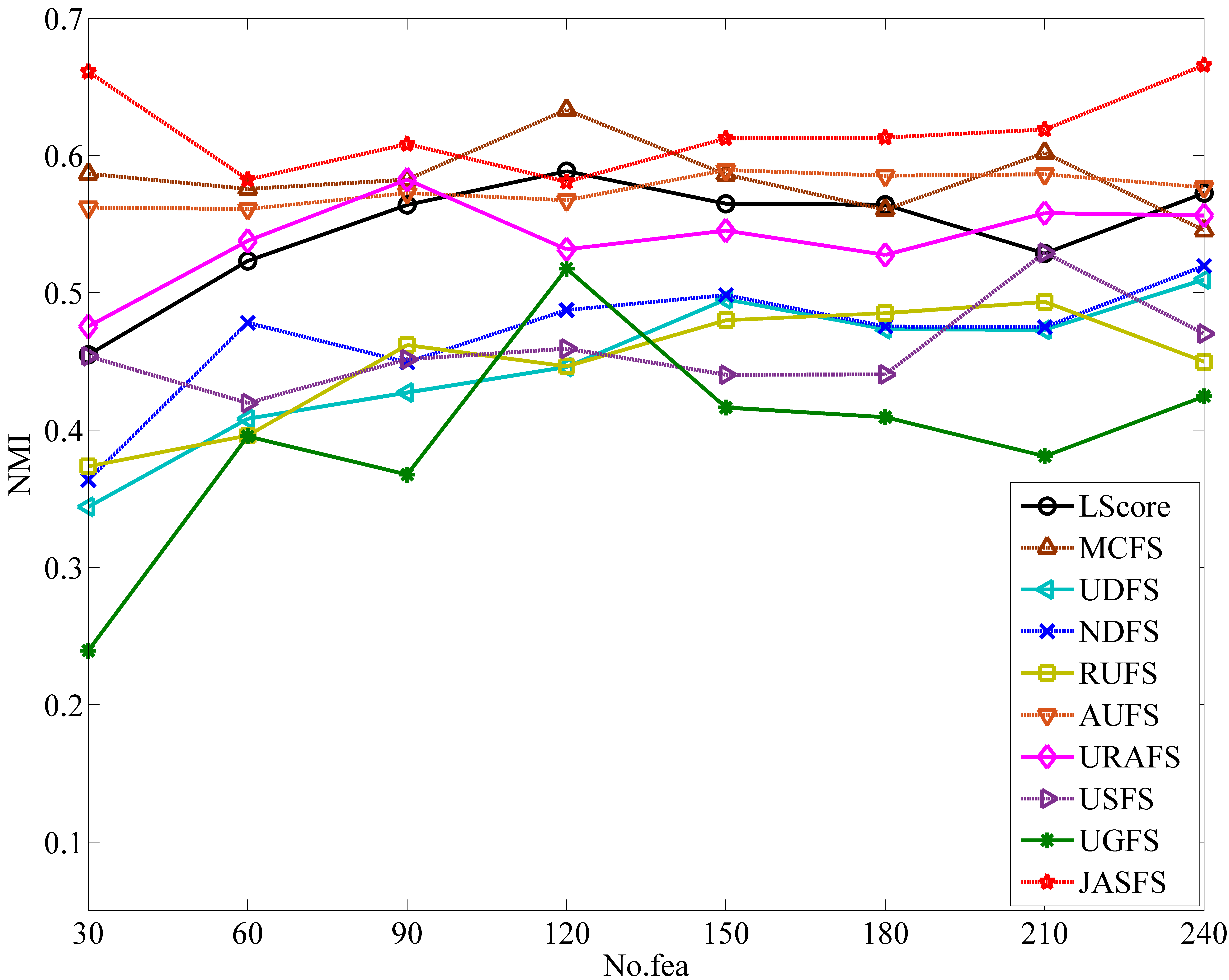}
    }
    \subfigure[]{\label{Fig:Breast3}
        \includegraphics[width=4cm,height=3.5cm]{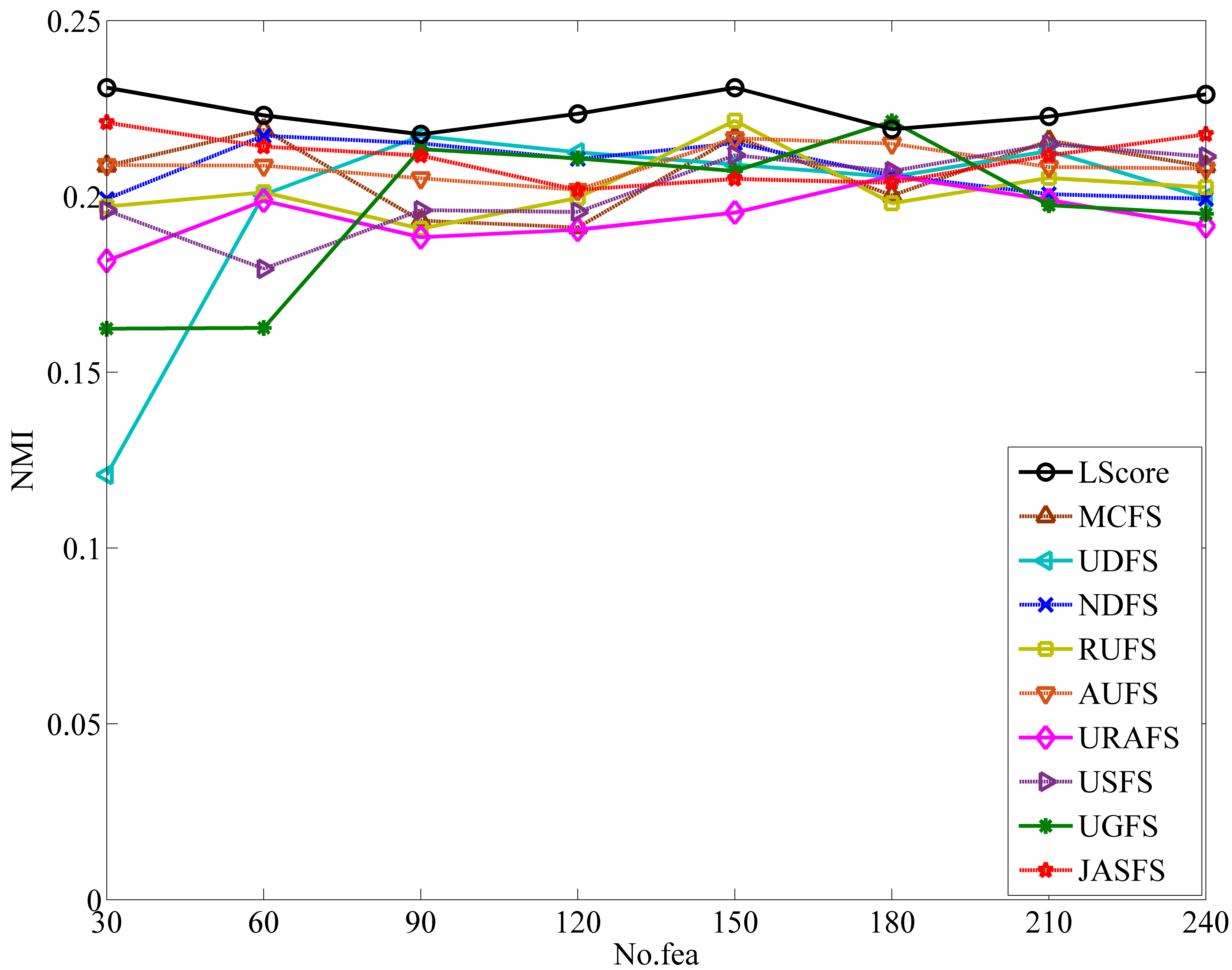}
    }
    \subfigure[]{\label{Fig:Jaffe}
        \includegraphics[width=4cm,height=3.5cm]{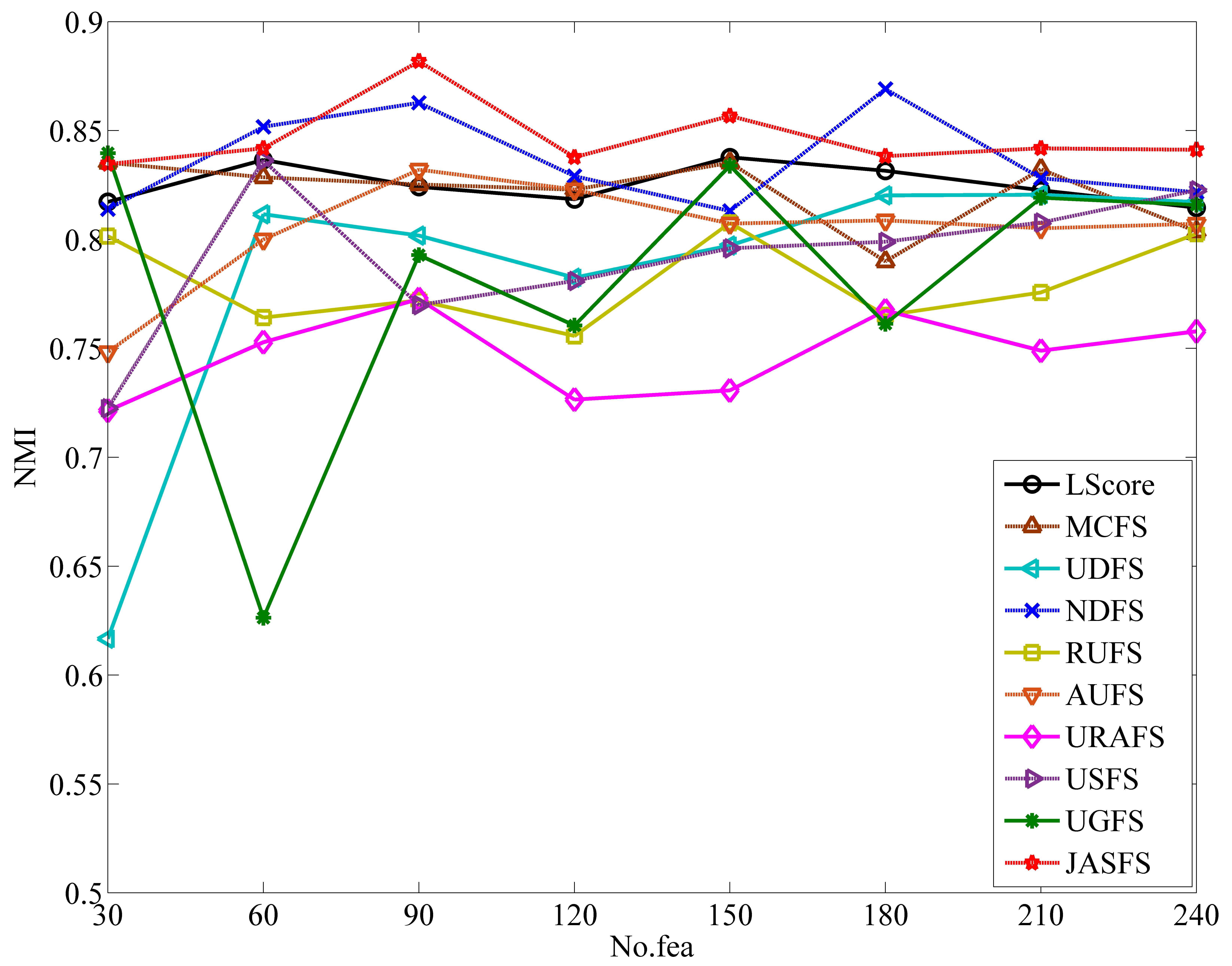}
    }
    \subfigure[]{\label{Fig:Lung}
        \includegraphics[width=4cm,height=3.5cm]{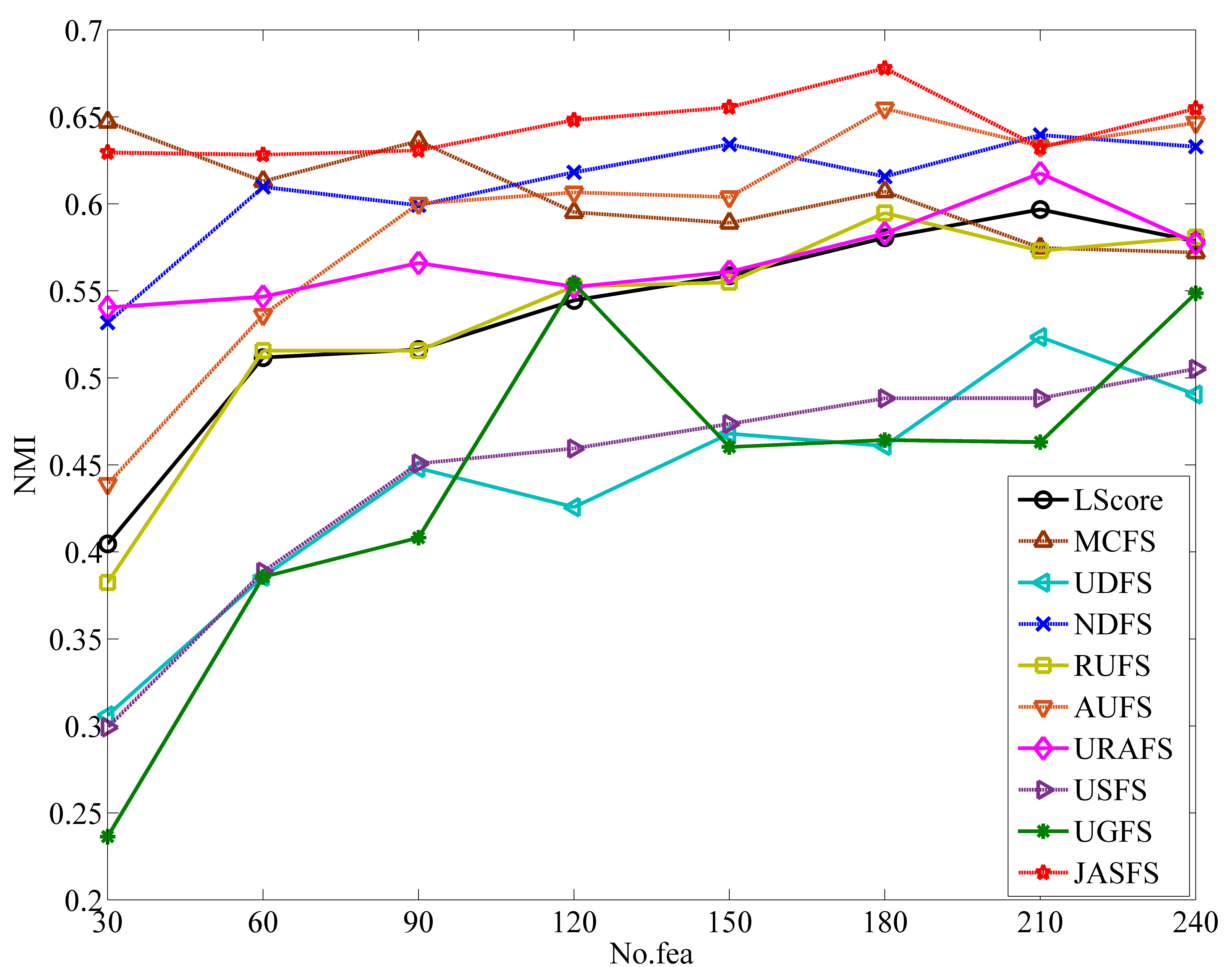}
    }
    \subfigure[]{\label{Fig:Mnist}
        \includegraphics[width=4cm,height=3.5cm]{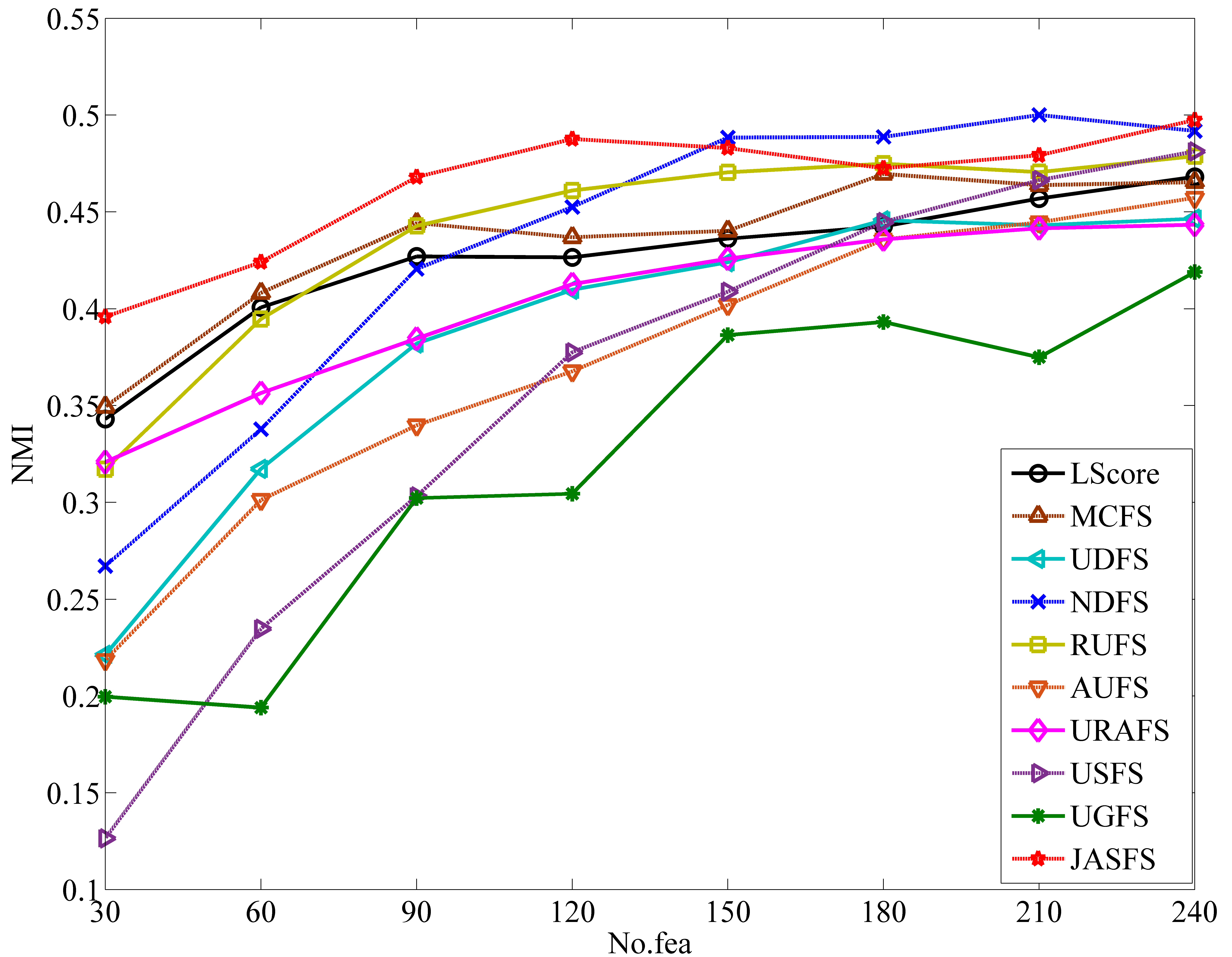}
    }
    \subfigure[]{\label{Fig:NCI}
        \includegraphics[width=4cm,height=3.5cm]{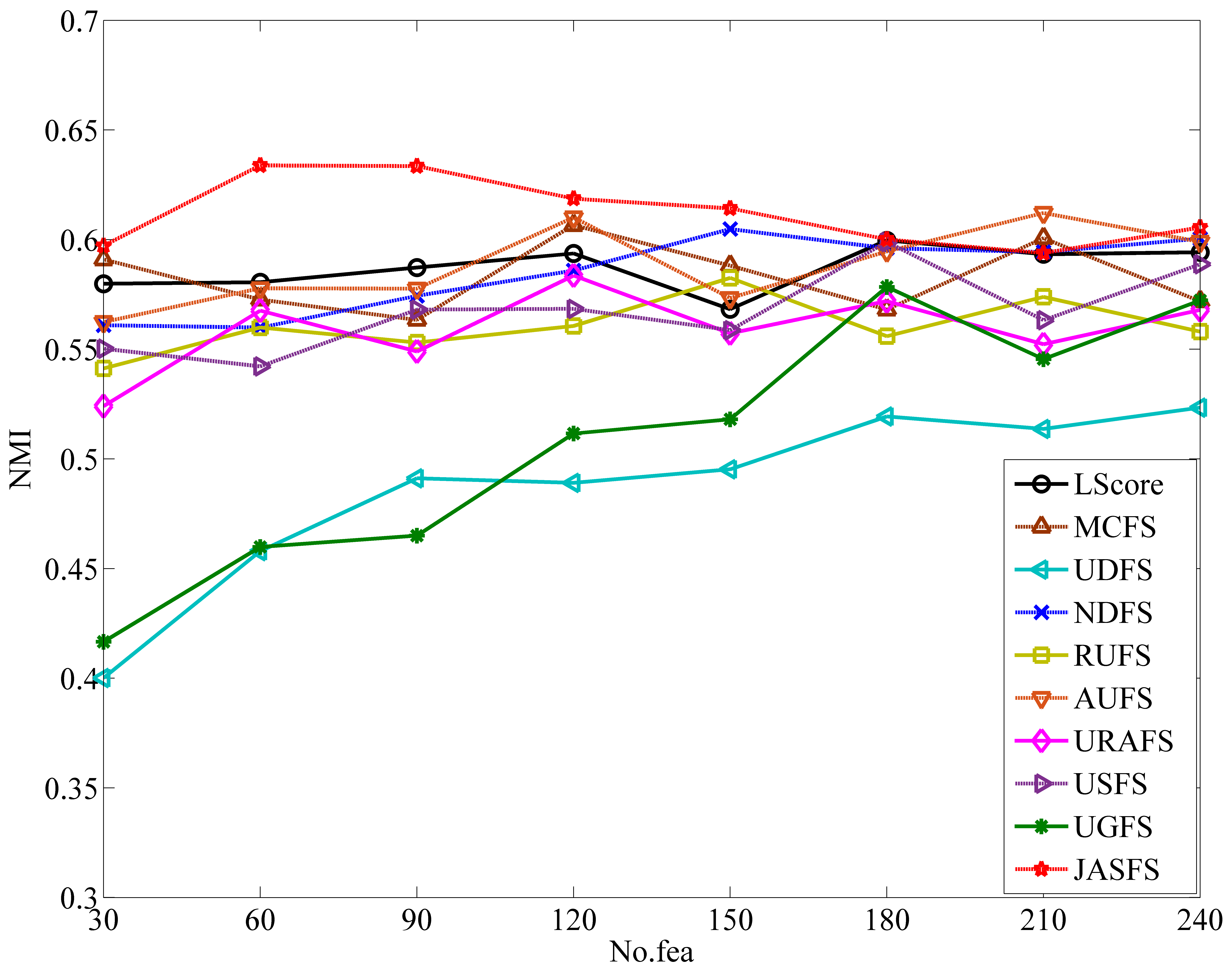}
    }
    \subfigure[]{\label{Fig:ORL}
        \includegraphics[width=4cm,height=3.5cm]{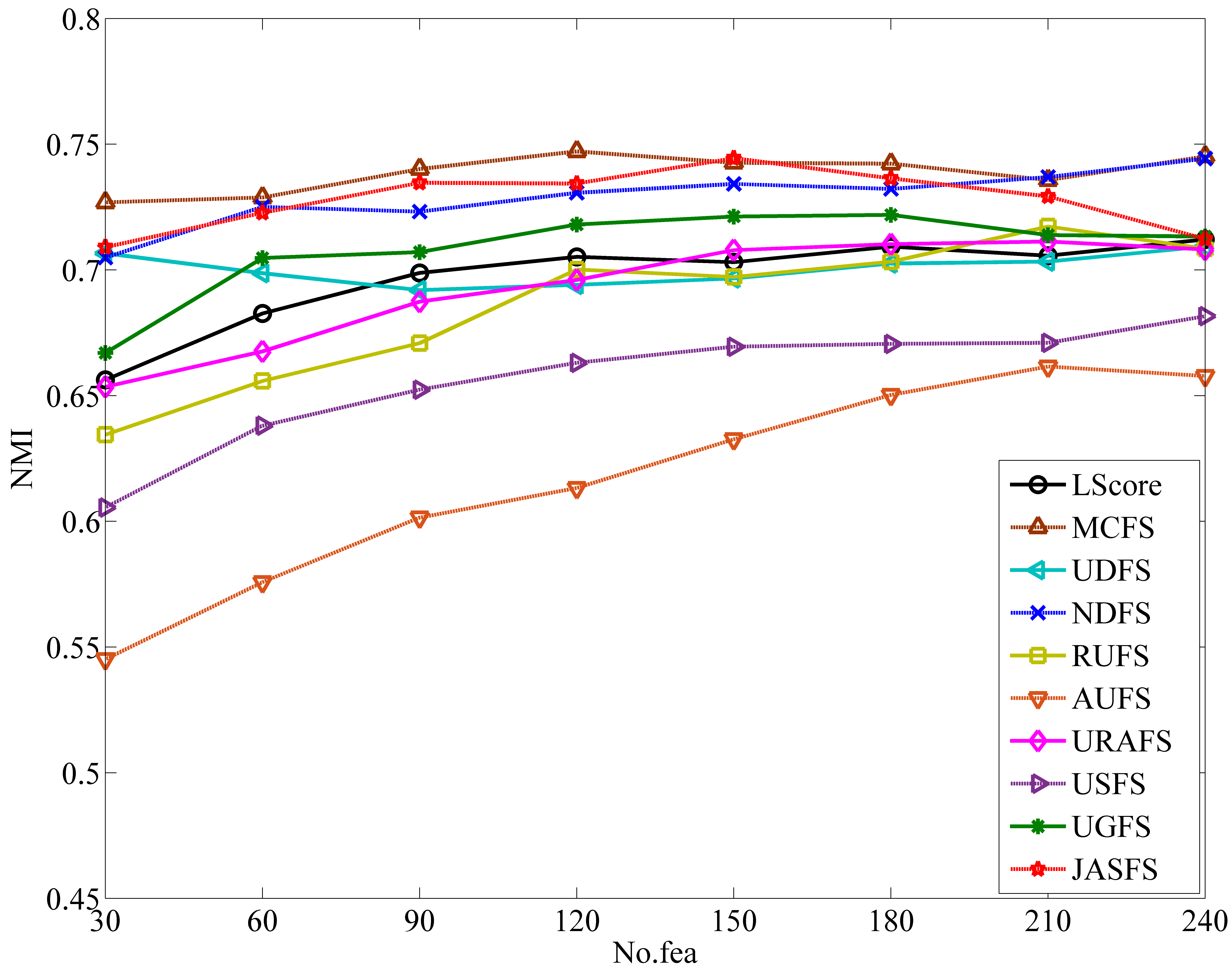}
    }
    \subfigure[]{\label{Fig:Palm}
        \includegraphics[width=4cm,height=3.5cm]{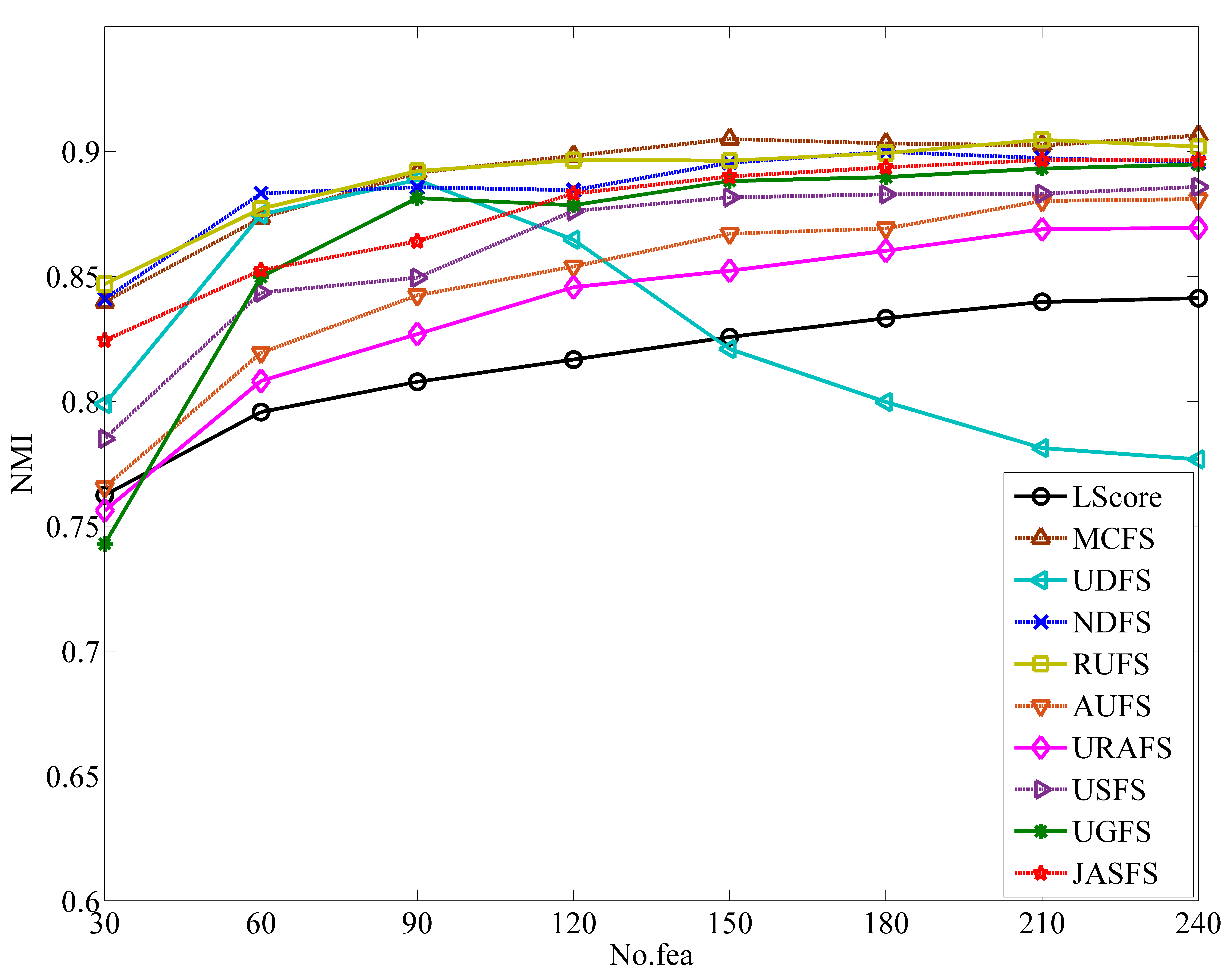}
    }
    \caption{\label{fig:nmivsnum}The NMI results with respect to different numbers of selected features.
      \subref{Fig:Brain}: Brain.
      \subref{Fig:Breast3}: Breast3.
      \subref{Fig:Jaffe}: Jaffe.
      \subref{Fig:Lung}: Lung.
      \subref{Fig:Mnist}: Mnist.
      \subref{Fig:NCI}: NCI.
      \subref{Fig:ORL}: ORL.
      \subref{Fig:Palm}: Palm.
      }
  \end{center}
\end{figure}

Second, we explore the influence of the number of selected features to the clustering results to measure the performance of each method. The No.fea ranges from $30$ to $240$ with the interval of $30$, and Figure~\ref{fig:accvsnum} and Figure \ref{fig:nmivsnum} show the ACC and NMI results, respectively. It is obvious that when No.fea is small ($30$), the clustering results of our method can beat other compared methods consistently in both ACC and NMI. When No.fea increases, our approach still obtain better or comparable results than other methods on most datasets. This clustering results demonstrate that our method can select more informative features than other methods when having the same number of selected features.

\subsection{Comparison of Running Time}
\begin{figure}[htp]
  \begin{center}
    \subfigure[]{\label{Fig:Brain1}
        \includegraphics[width=4cm,height=3.5cm]{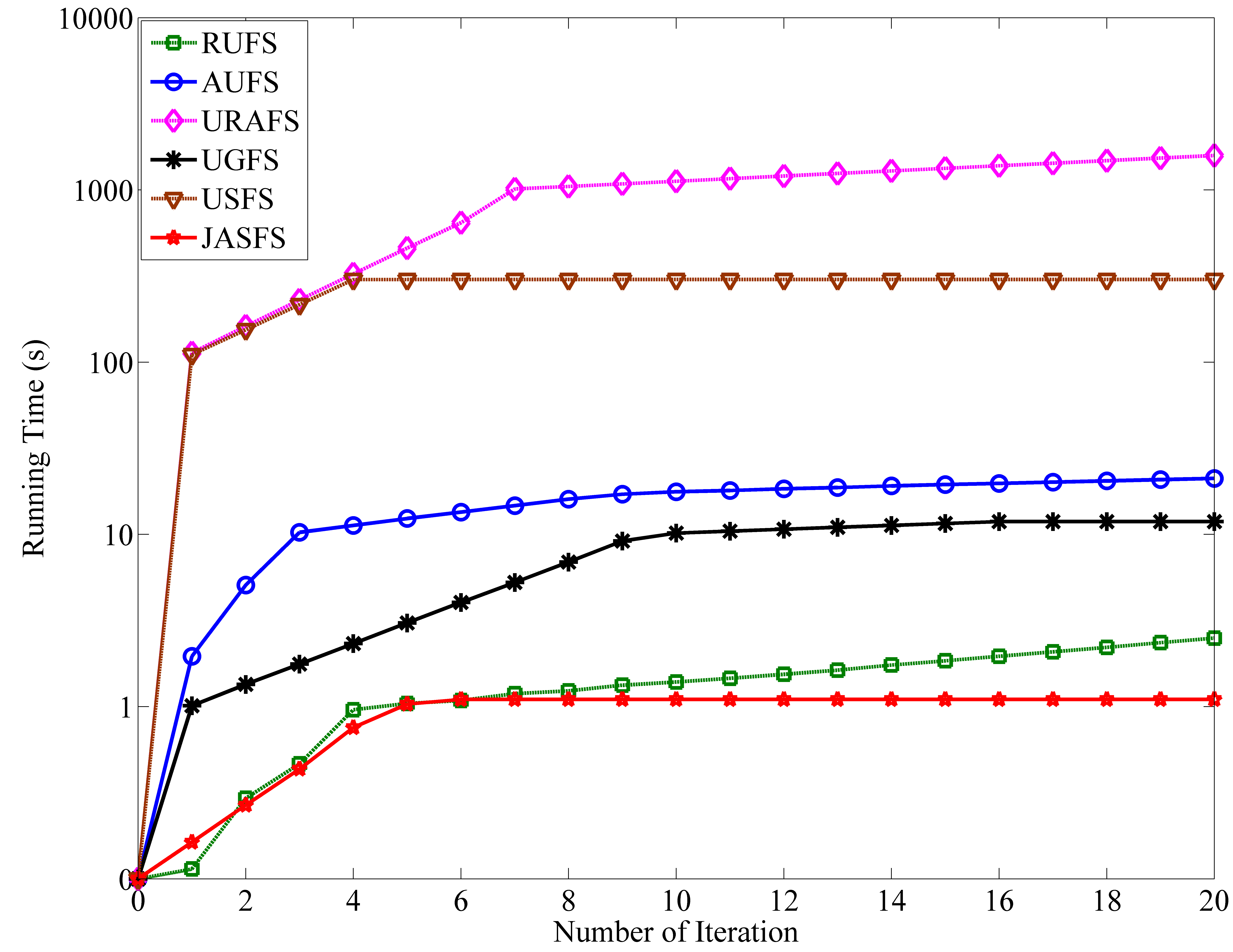}
    }
     \subfigure[]{\label{Fig:Breast31}
        \includegraphics[width=4cm,height=3.5cm]{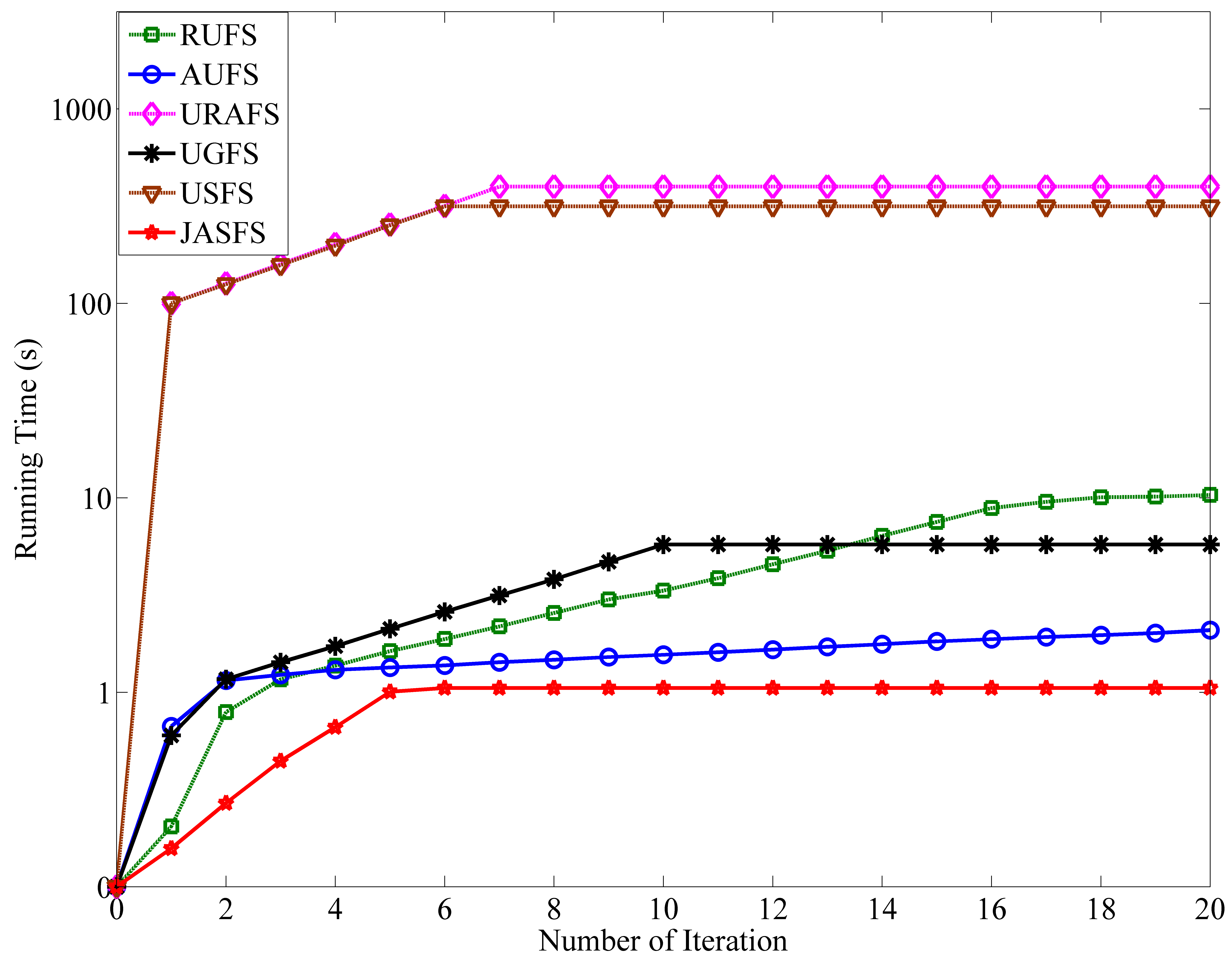}
    }
     \subfigure[]{\label{Fig:Jaffe1}
        \includegraphics[width=4cm,height=3.5cm]{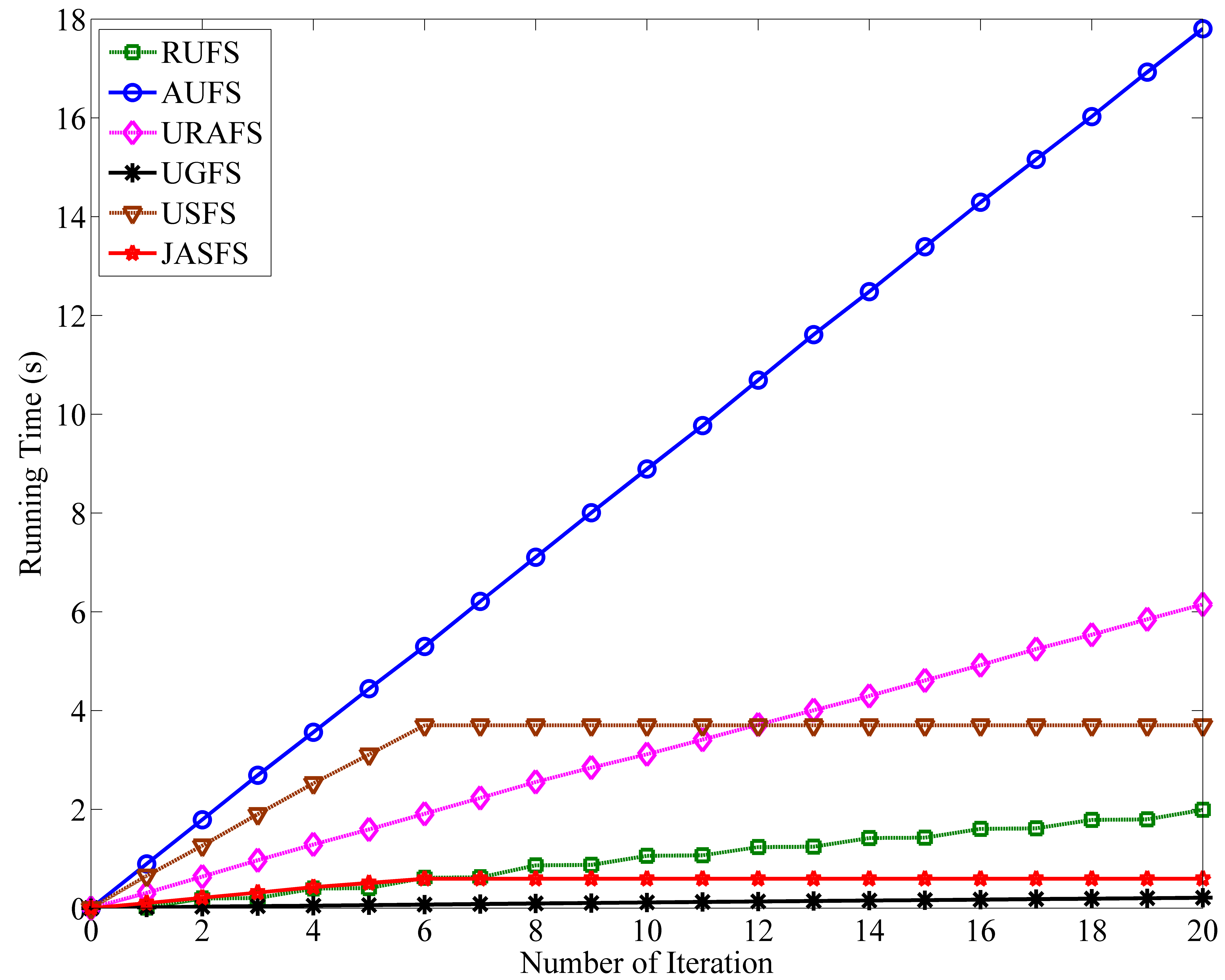}
    }
     \subfigure[]{\label{Fig:Lung1}
        \includegraphics[width=4cm,height=3.5cm]{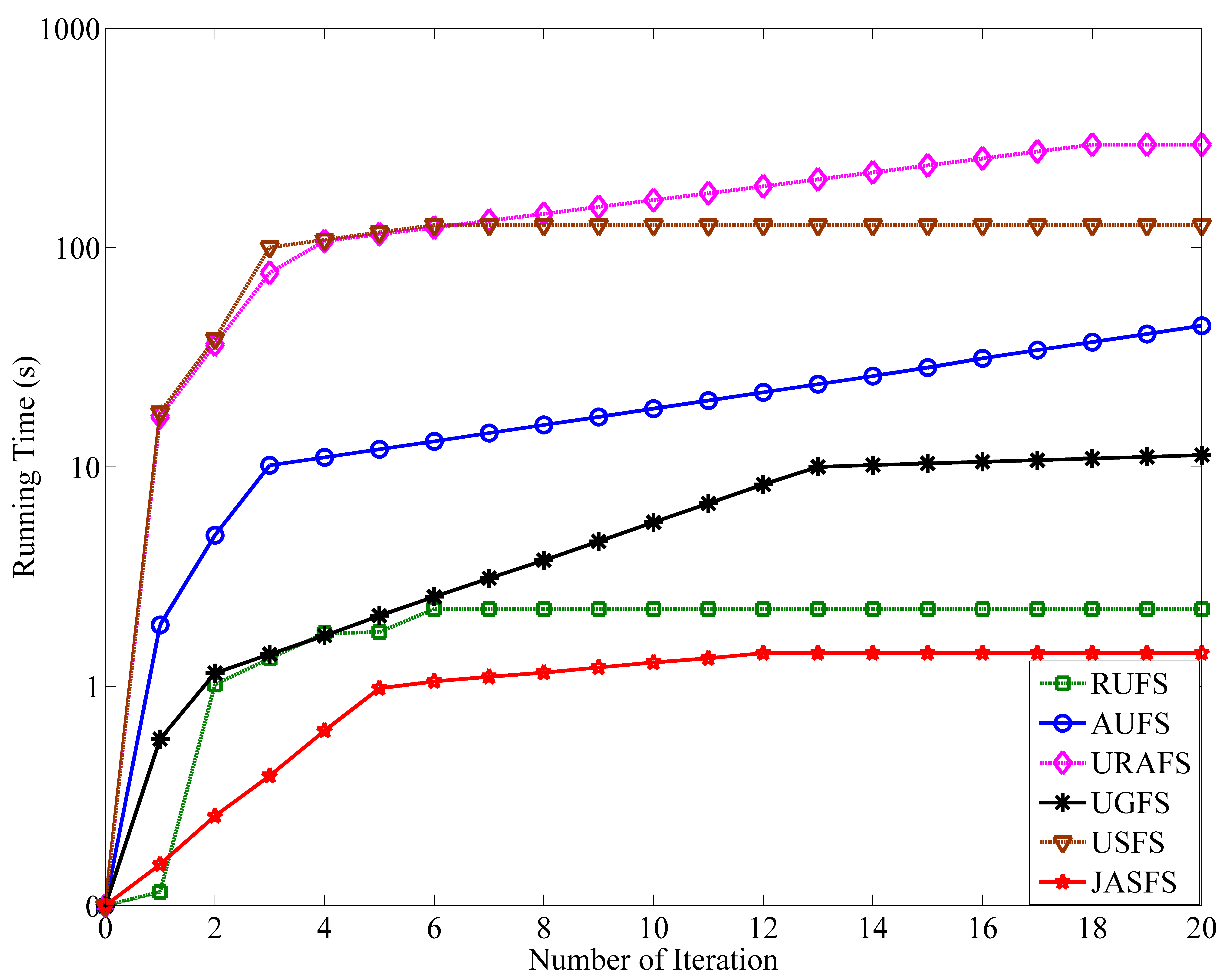}
    }
     \subfigure[]{\label{Fig:Mnist1}
        \includegraphics[width=4cm,height=3.5cm]{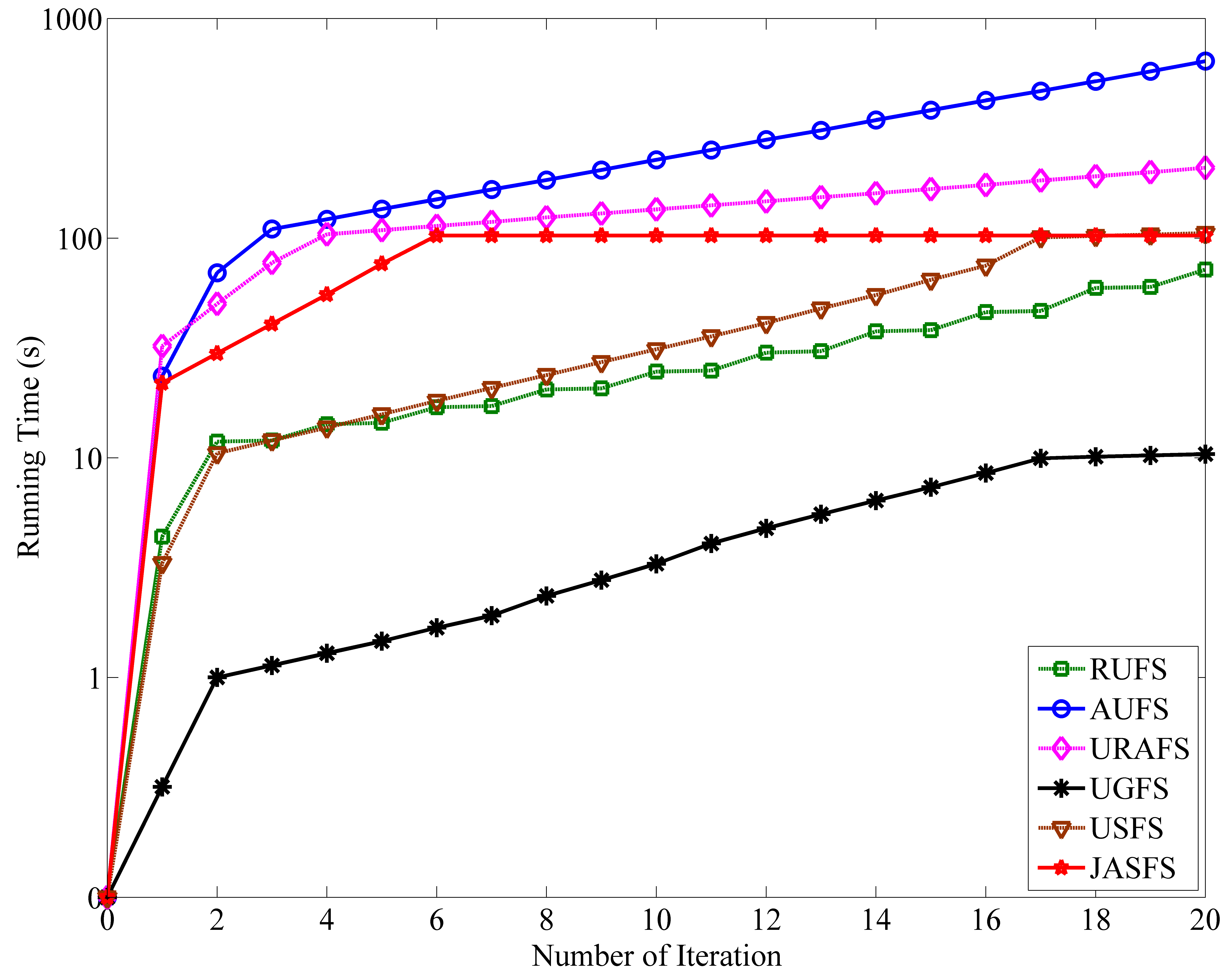}
    }
    \subfigure[]{\label{Fig:NCI1}
        \includegraphics[width=4cm,height=3.5cm]{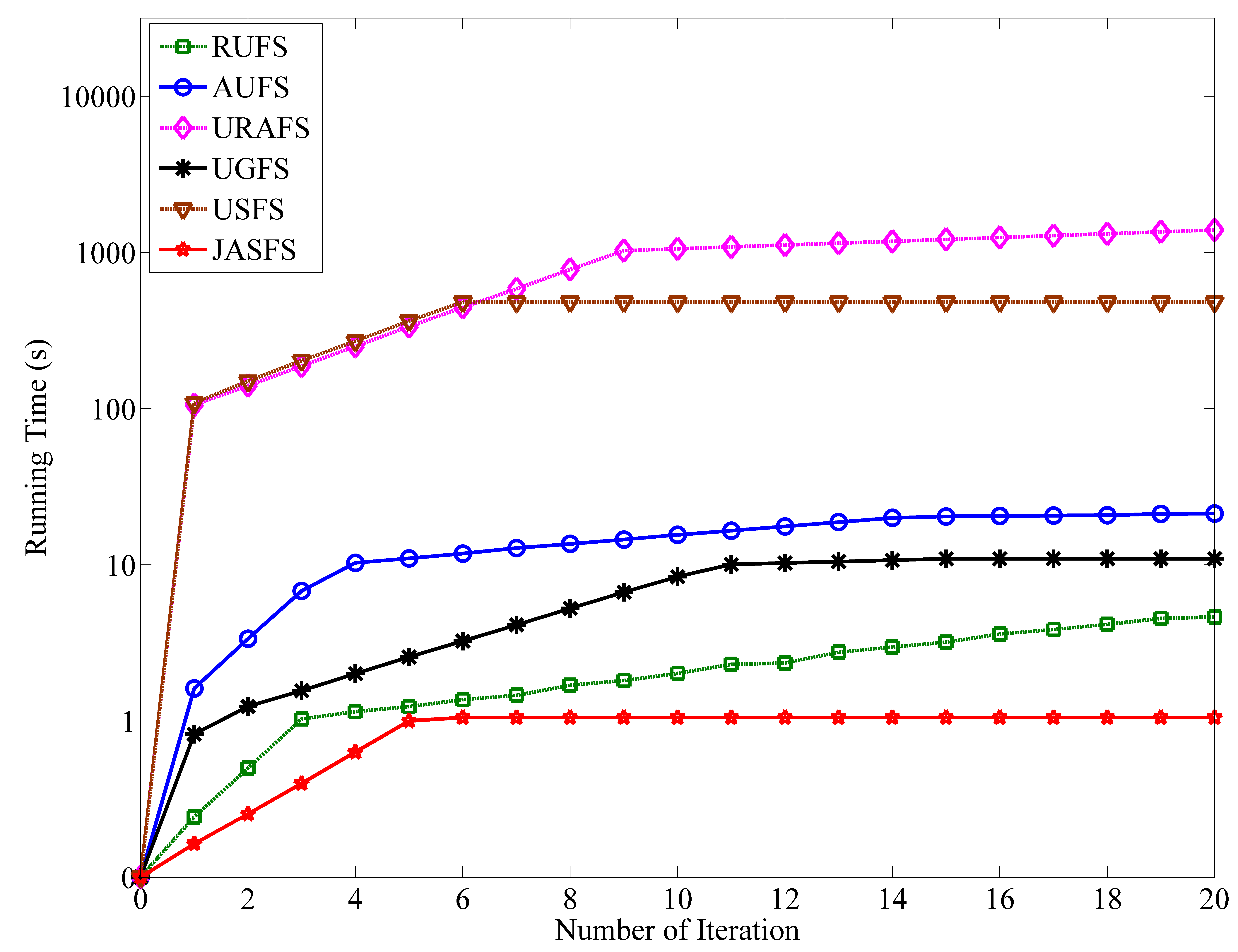}
    }
     \subfigure[]{\label{Fig:ORL1}
        \includegraphics[width=4cm,height=3.5cm]{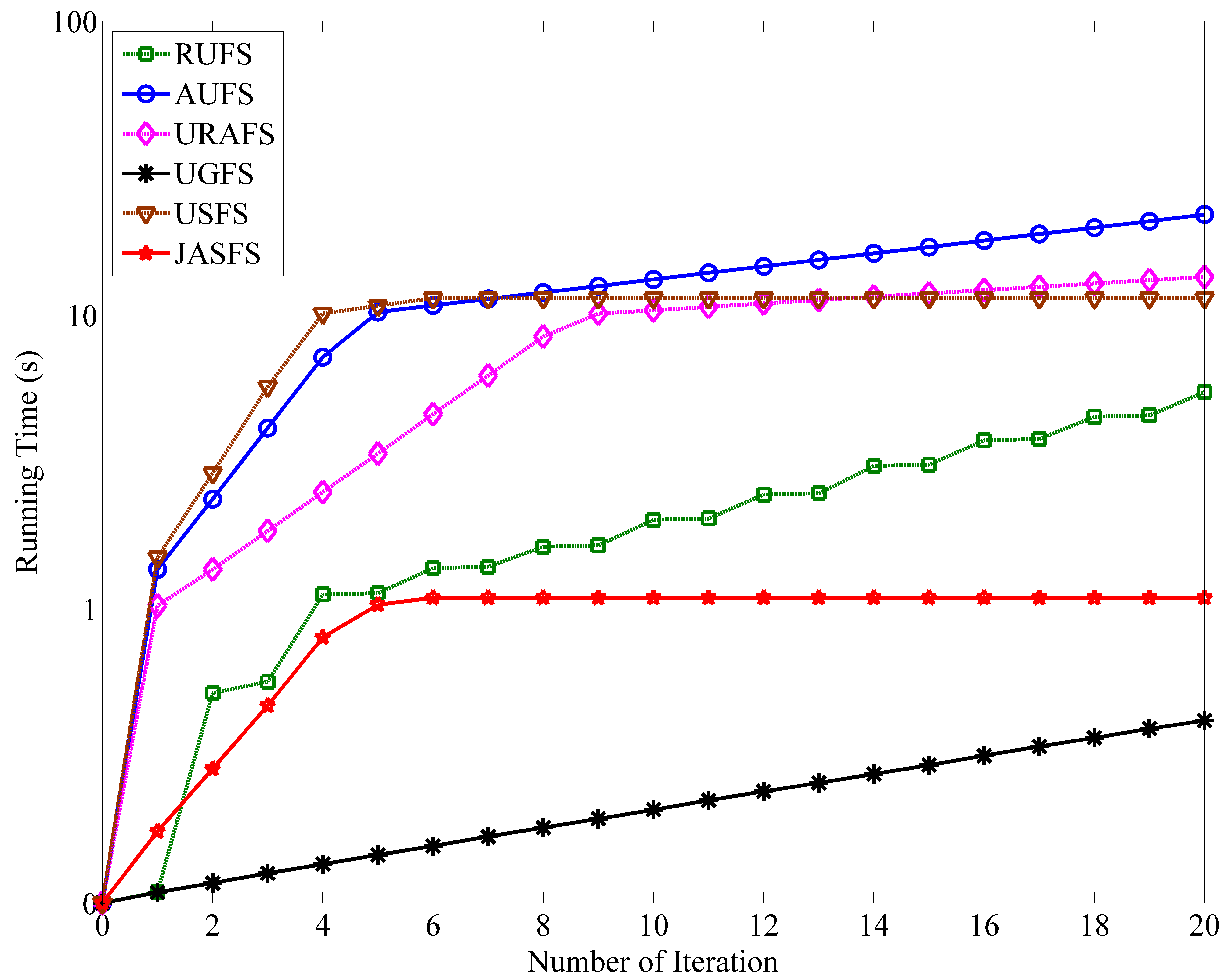}
    }
     \subfigure[]{\label{Fig:Palm1}
        \includegraphics[width=4cm,height=3.5cm]{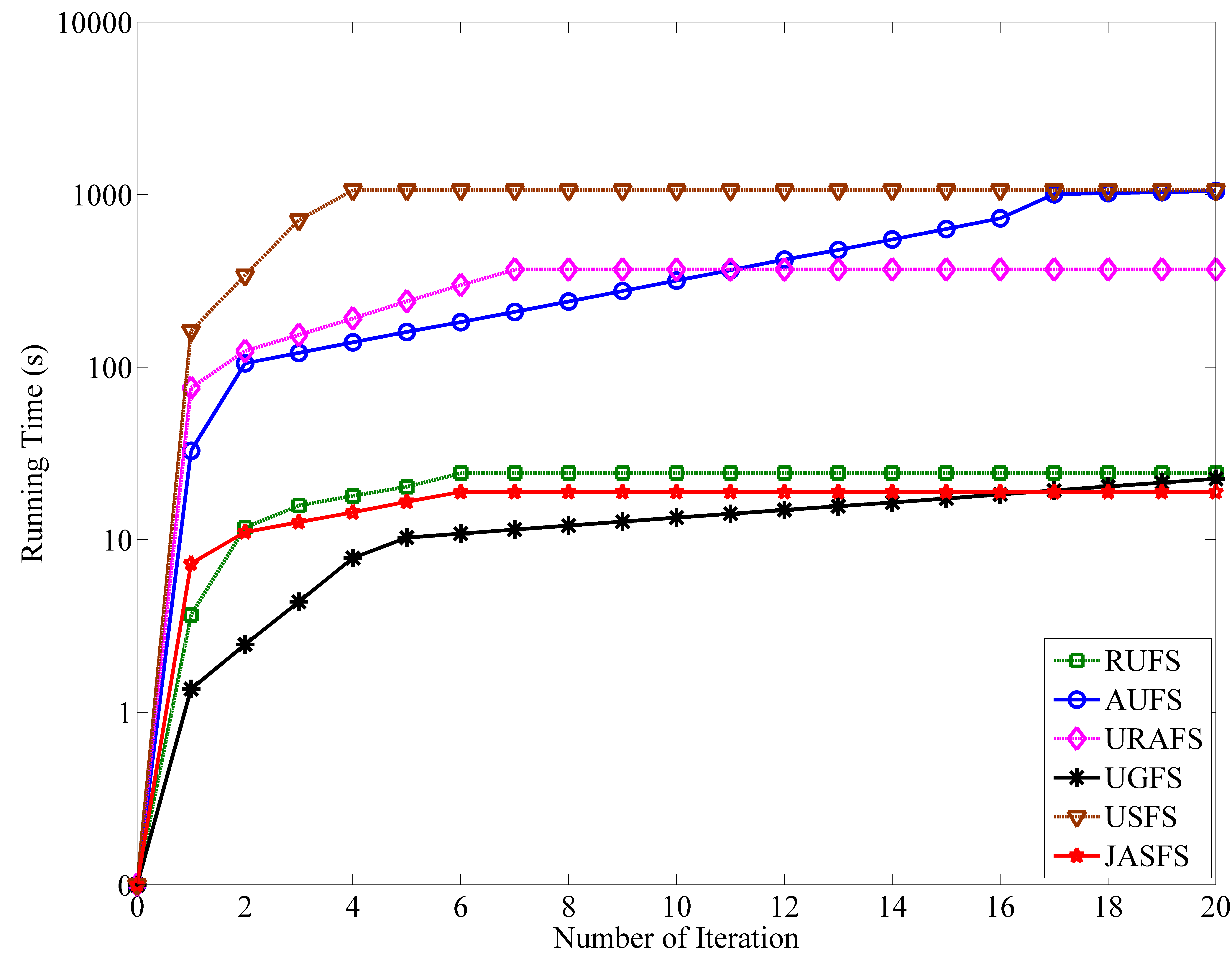}
    }
    \caption{\label{fig:runningtime}The comparison of running time.
      \subref{Fig:Brain1}: Brain.
      \subref{Fig:Breast31}: Breast3.
      \subref{Fig:Jaffe1}: Jaffe.
      \subref{Fig:Lung1}: Lung.
      \subref{Fig:Mnist1}: Mnist.
      \subref{Fig:NCI1}: NCI.
      \subref{Fig:ORL1}: ORL.
      \subref{Fig:Palm1}: Palm.
      }
  \end{center}
\end{figure}

\begin{figure}[htp]
  \begin{center}
    \subfigure[]{\label{Fig:obj-breast3}
        \includegraphics[width=4cm,height=3.5cm]{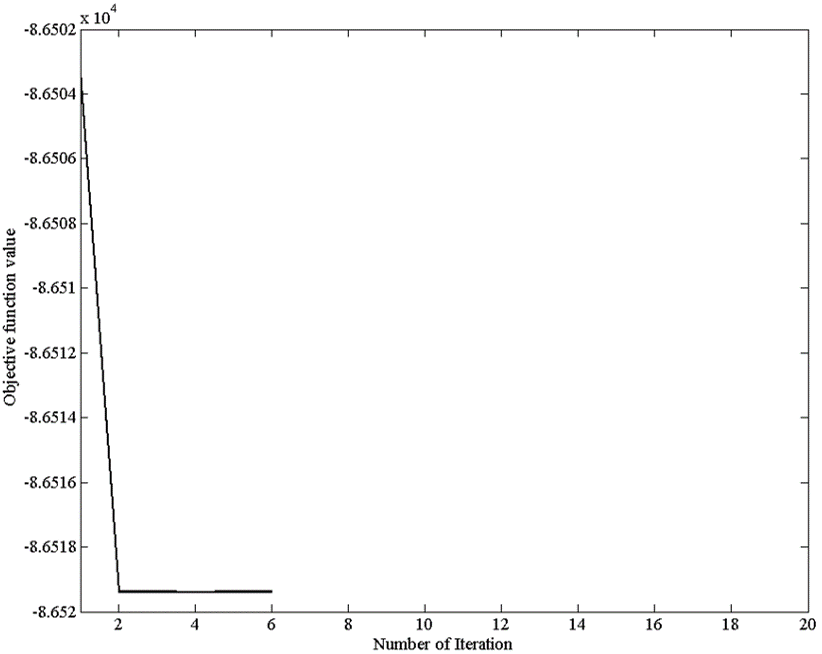}
    }
     \subfigure[]{\label{Fig:obj-lung}
        \includegraphics[width=4cm,height=3.5cm]{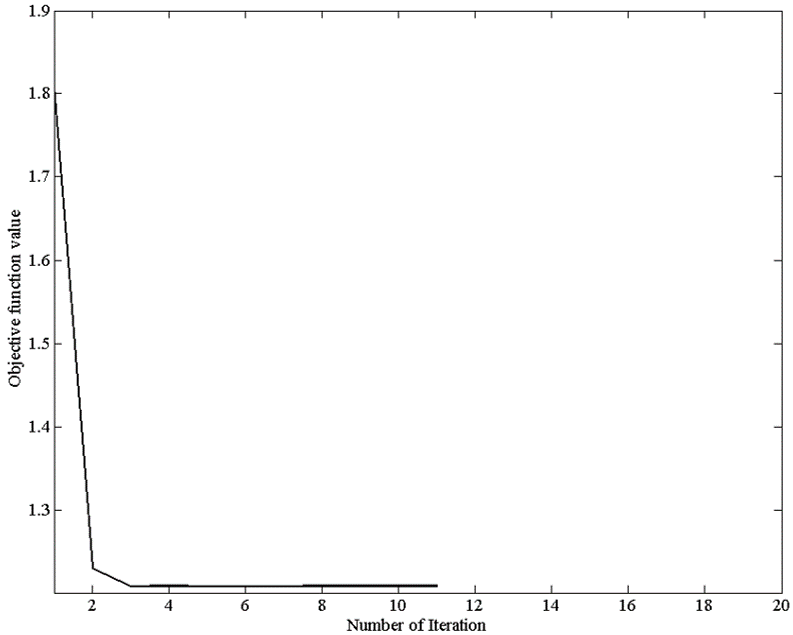}
    }
    \caption{\label{fig:obj}Objective value curves of JASFS on:
      \subref{Fig:obj-breast3}: Breast3.
      \subref{Fig:obj-lung}: Lung.
      }
  \end{center}
\end{figure}

In this subsection, we conduct experiment to show the running time of some compared methods on all datasets. In this experiment, the max number of iteration is set as $20$, and the convergence tolerance is set as $10^{-5}$, Figure~\ref{fig:runningtime} shows the results. It can be seen from this figure that, the running time of each method level off to a value after several iterations, which accounts for the convergence of each method. As shows in this figure, the running time of our approach is much less than the comparison methods on most datasets, since the proposed JASFS method converges first and has lower computational complexity when data feature is high-dimensional. Figure \ref{fig:obj} shows two examples of the objective function value curve. With this figure we can see that the objective function value decreases sharply at first and does not change significantly after several iterations, which demonstrates that our proposed optimization method is able to converge efficiently.

\subsection{Parameter Sensitivity}
\begin{figure}[htp]
  \begin{center}
    \subfigure[]{\label{Fig:ACC-NCI1}
        \includegraphics[width=4.5cm,height=3.7cm]{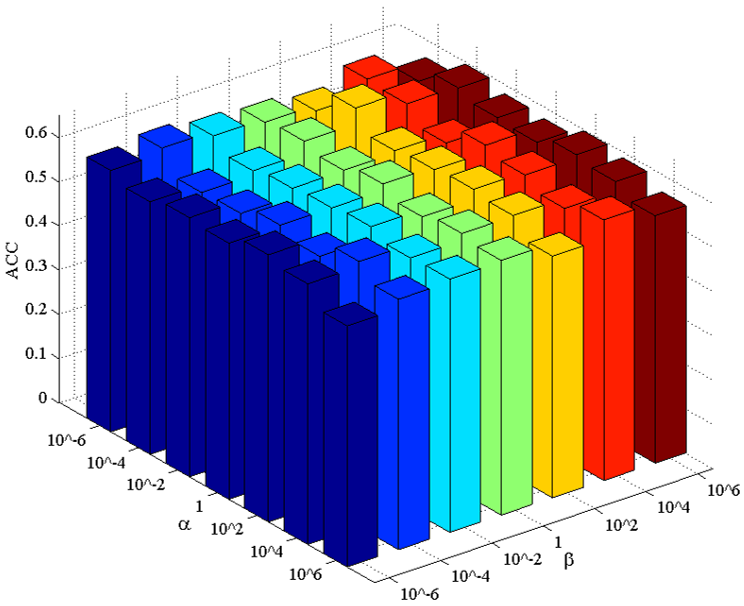}
    }
    \subfigure[]{\label{Fig:ACC-NCI2}
        \includegraphics[width=4.5cm,height=3.7cm]{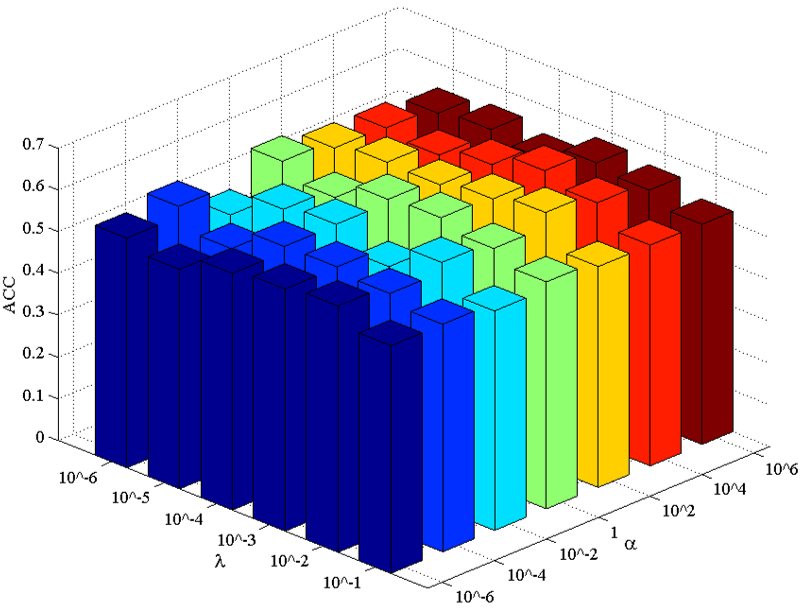}
    }
    \subfigure[]{\label{Fig:ACC-NCI3}
        \includegraphics[width=4.5cm,height=3.7cm]{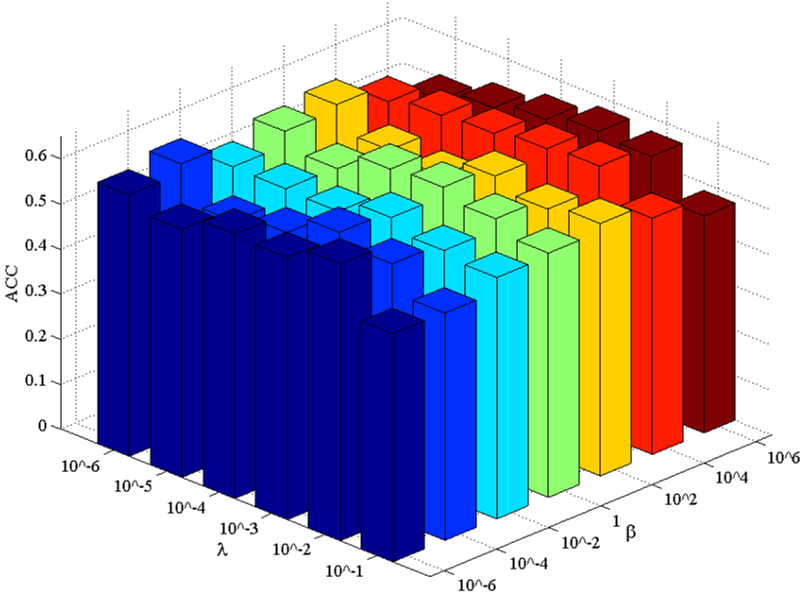}
    }
    \subfigure[]{\label{Fig:NMI-NCI1}
        \includegraphics[width=4.5cm,height=3.7cm]{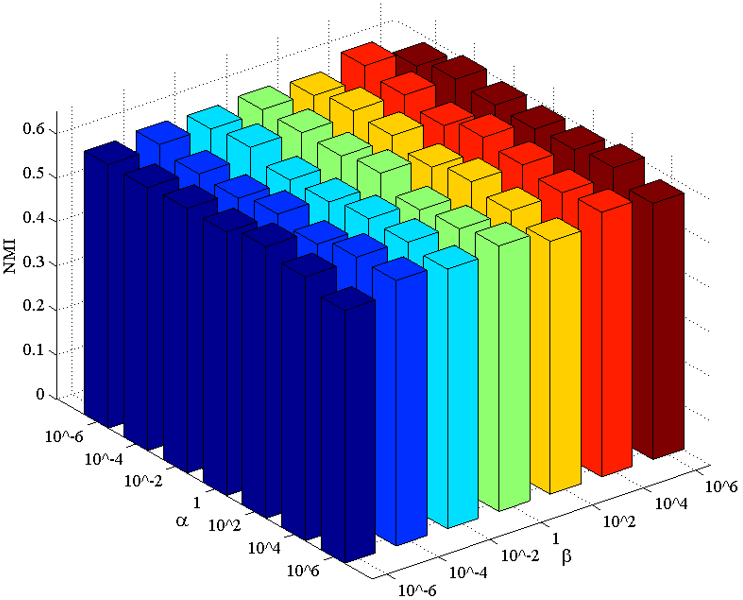}
    }
    \subfigure[]{\label{Fig:NMI-NCI2}
        \includegraphics[width=4.5cm,height=3.7cm]{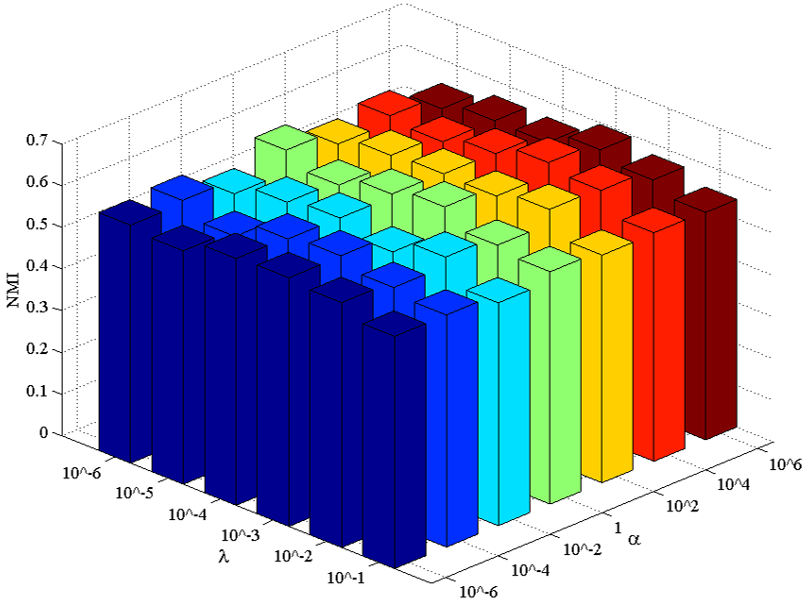}
    }
    \subfigure[]{\label{Fig:NMI-NCI3}
        \includegraphics[width=4.5cm,height=3.7cm]{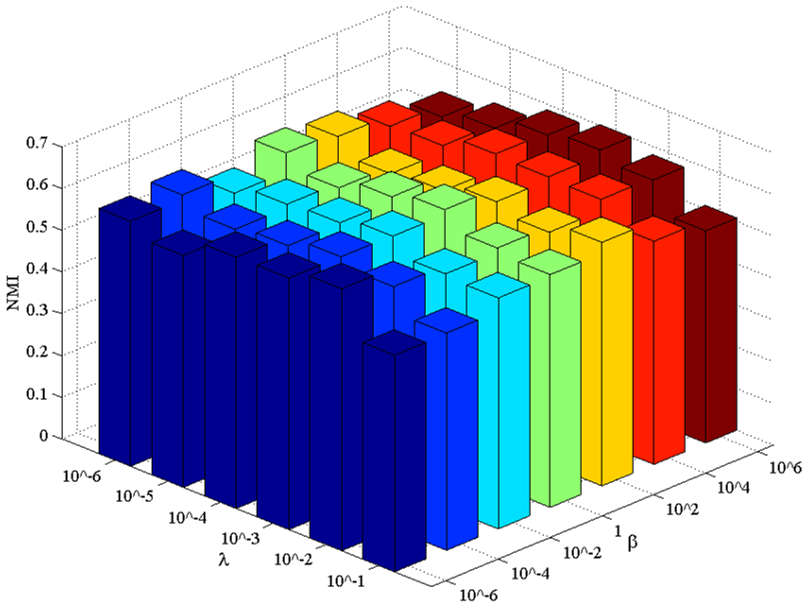}
    }
    \caption{\label{fig:parametersonnci}Parameter sensitivity demonstration on NCI data set. Top: results of ACC; Bottom: results of NMI.
      \subref{Fig:ACC-NCI1} and \subref{Fig:NMI-NCI1}: $\lambda = 1e-3$;
      \subref{Fig:ACC-NCI2} and \subref{Fig:NMI-NCI2}: $\beta =1$;
      \subref{Fig:ACC-NCI3} and \subref{Fig:NMI-NCI3}: $\alpha = 1$.
      }
  \end{center}
\end{figure}

\begin{figure}[htp]
  \begin{center}
    \subfigure[]{\label{Fig:ACC-ORL1}
        \includegraphics[width=4.5cm,height=3.7cm]{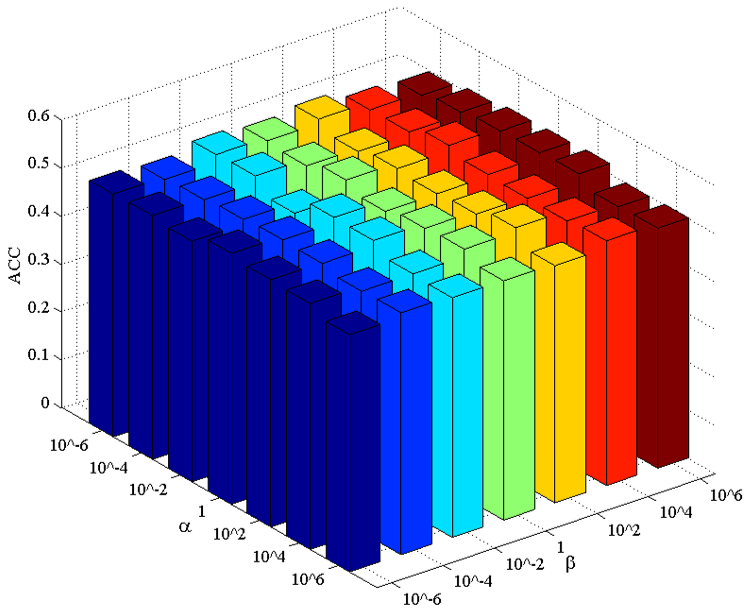}
    }
    \subfigure[]{\label{Fig:ACC-ORL2}
        \includegraphics[width=4.5cm,height=3.7cm]{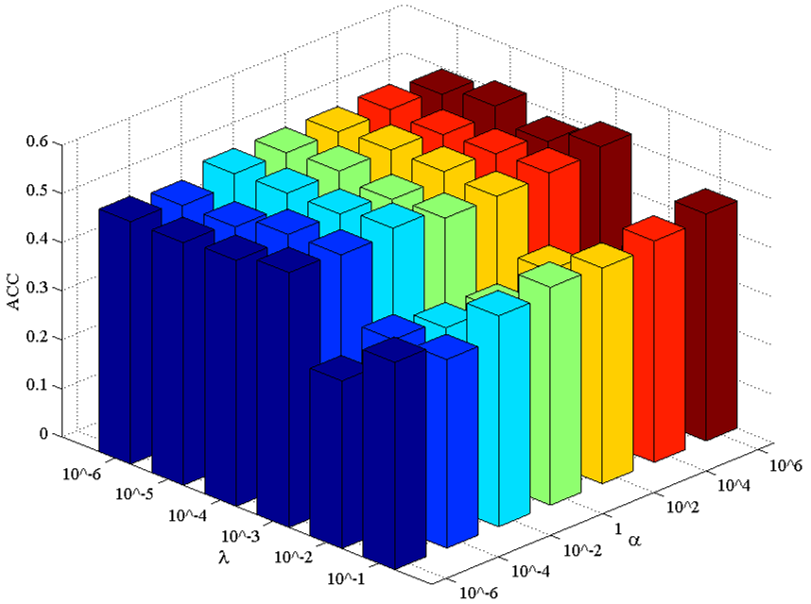}
    }
    \subfigure[]{\label{Fig:ACC-ORL3}
        \includegraphics[width=4.5cm,height=3.7cm]{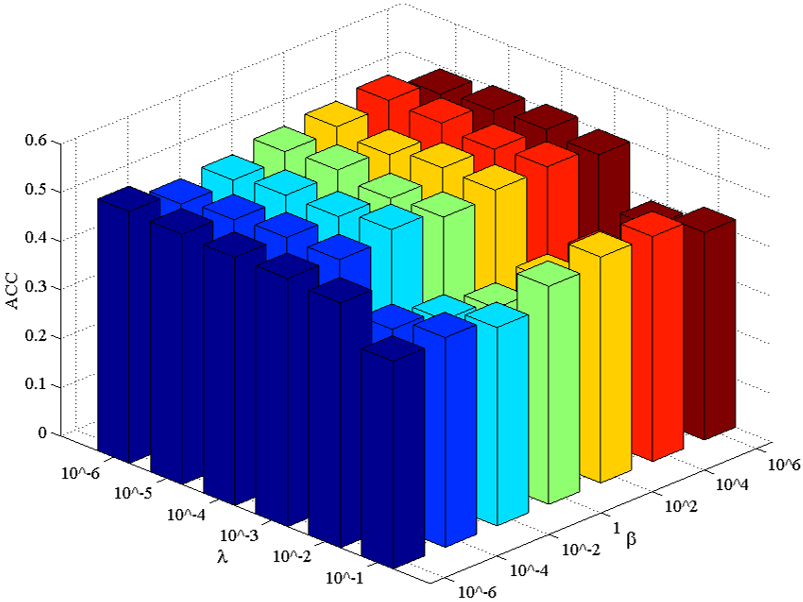}
    }
    \subfigure[]{\label{Fig:NMI-ORL1}
        \includegraphics[width=4.5cm,height=3.7cm]{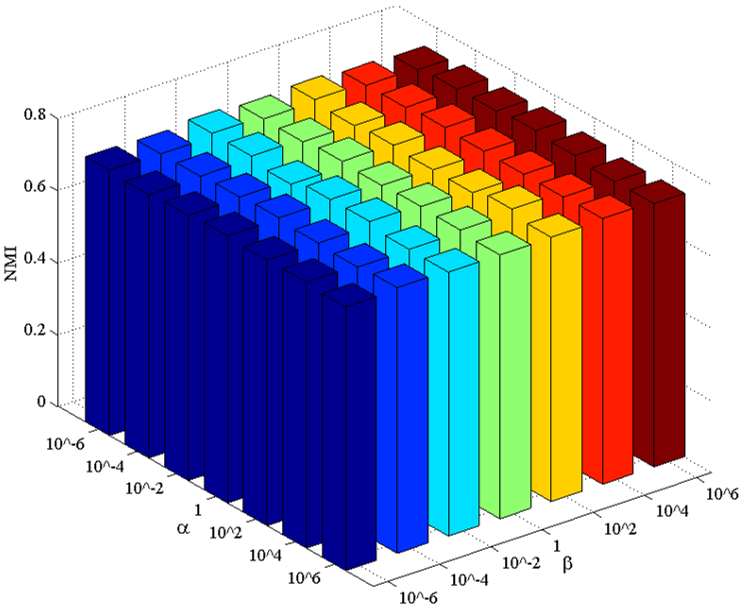}
    }
    \subfigure[]{\label{Fig:NMI-ORL2}
        \includegraphics[width=4.5cm,height=3.7cm]{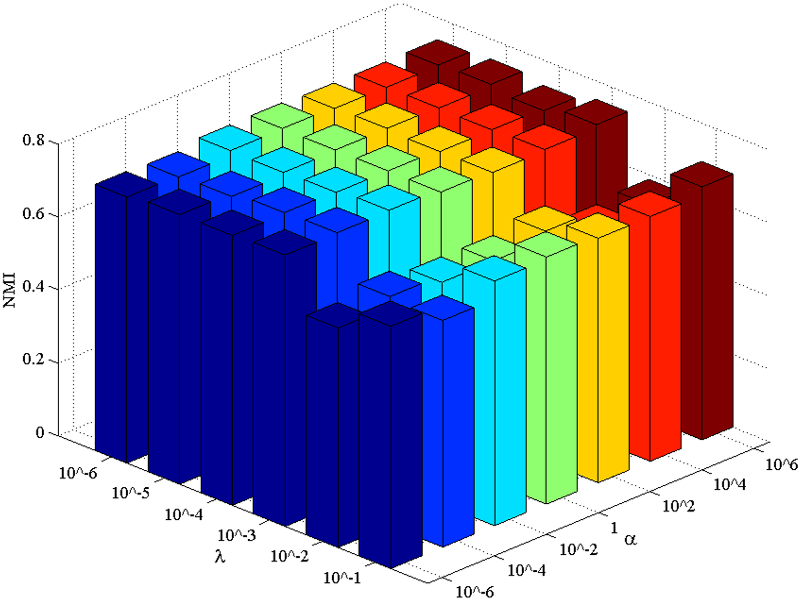}
    }
    \subfigure[]{\label{Fig:NMI-ORL3}
        \includegraphics[width=4.5cm,height=3.7cm]{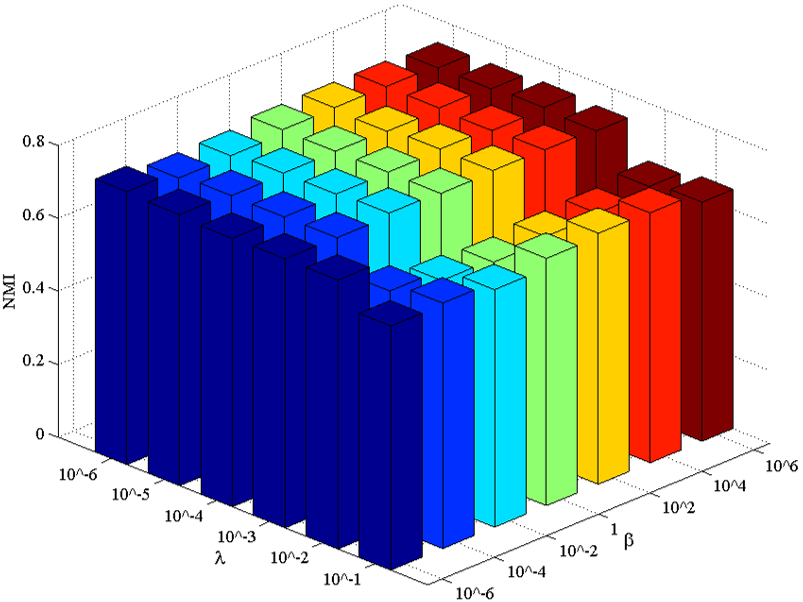}
    }
    \caption{\label{fig:parametersonorl}Parameter sensitivity demonstration on ORL data set. Top: results of ACC; Bottom: results of NMI.
      \subref{Fig:ACC-ORL1} and \subref{Fig:NMI-ORL1}: $\lambda = 1e-3$;
      \subref{Fig:ACC-ORL2} and \subref{Fig:NMI-ORL2}: $\beta =1$;
      \subref{Fig:ACC-ORL3} and \subref{Fig:NMI-ORL3}: $\alpha = 1$.
      }
  \end{center}
\end{figure}

In this section, we investigate the sensitivity of parameters $\lambda$, $\alpha$, and $\beta$ in JASFS. Due to the space limit, we only report the results on NCI and ORL, which are shown in Figure~\ref{fig:parametersonnci} and Figure~\ref{fig:parametersonorl} respectively. It can be seen that the performance of JASFS is hardly sensitive to $\alpha$ and $\beta$, while is little sensitive to $\lambda$, especially when the value of $\lambda$ is large (more than $10^{-2}$). The reason is that $\lambda$ will influence the number of selected features, when $\lambda$ is too large, JASFS will obtain a very sparse solution and the number of selected features tends to be zero. While when $\lambda$ is in the range of $[10^{-6},10^{-3}]$, JASFS is not very sensitive to it. The results show that our method is robust to the parameters $\lambda$, $\alpha$, and $\beta$.

\section{Conclusion and Future Work}
\label{Sec:Conclusion}

We have proposed a novel unsupervised feature selection method, called JASFS, in which nonnegative spectral clustering, adaptive graph learning and $l_{2,0}$-norm minimization are performed simultaneously. To solve the proposed objective function, an efficient alternative algorithm is derived, in which an accelerated matrix homotopy iterative hard thresholding (AMHIHT) method is proposed to optimize the transformation matrix. After optimization, a group of optimal features will be selected according to the corresponding non-zero rows of transformation matrix, and the number of selected features can be determined automatically. Experiment results on eight benchmark datasets show that our approach outperforms current state-of-the-art unsupervised feature selection methods evaluated in terms of cluster accuracy and normalized mutual information. In the proposed objective function, there are three hyper-parameters, which are set empirically and tuned experimentally, it will consume a lot of time and labor. How to select the optimal hyper-parameters adaptively is an interesting and challenging problem that deserves to investigate in the future.

\section*{Acknowledgments}
This work was supported in part by National Natural Science Foundation of China (NSFC) under grant \#61873067, in part by the the Scientific Research Funds of Huaqiao University under grant \#21BS121.

\bibliographystyle{spbasic}
\bibliography{references}
\end{document}